\documentclass[runningheads]{llncs}
\usepackage{graphicx}
\usepackage{comment}
\usepackage{amsmath,amssymb} 
\usepackage{color}

\def\finalversion{1}

\ifx\finalversion\undefined
\usepackage{ruler}
\usepackage[width=122mm,left=12mm,paperwidth=146mm,height=193mm,top=12mm,paperheight=217mm]{geometry}
\fi
\usepackage[pagebackref=true,breaklinks=true,colorlinks=true,linkcolor=red,citecolor=blue,urlcolor=magenta]{hyperref}

\usepackage{commath,dsfont,complexity,mathtools,nicefrac,booktabs}
\usepackage[linesnumbered,ruled,noend]{algorithm2e}

\def\ddefloop#1{\ifx\ddefloop#1\else\ddef{#1}\expandafter\ddefloop\fi}
\def\ddef#1{\expandafter\def\csname b#1\endcsname{\ensuremath{\mathbb{#1}}}}
\ddefloop ABCDEFGHIJKLMNOPQRSTUVWXYZ\ddefloop
\def\ddef#1{\expandafter\def\csname c#1\endcsname{\ensuremath{\mathcal{#1}}}}
\ddefloop ABCDEFGHIJKLMNOPQRSTUVWXYZ\ddefloop

\DeclareMathOperator{\softmax}{softmax}
\DeclareMathOperator{\sigmoid}{sigmoid}

\def\paranoskip#1{\noindent\textbf{\boldmath #1}}
\def\para#1{\medskip\noindent\textbf{\boldmath #1}}

\begin{document}
\pagestyle{headings}
\mainmatter
\def\ECCVSubNumber{5495}

\title{Boosting Weakly Supervised Object Detection with Progressive Knowledge Transfer}

\ifx\finalversion\undefined
\titlerunning{ECCV-20 submission ID \ECCVSubNumber} 
\authorrunning{ECCV-20 submission ID \ECCVSubNumber} 
\author{Anonymous ECCV submission}
\institute{Paper ID \ECCVSubNumber}
\else
\titlerunning{WSOD with Progressive Knowledge Transfer}
\author{
Yuanyi Zhong\inst{1} \thanks{Part of this work was done when the first author was an intern at Microsoft.} \and
Jianfeng Wang\inst{2} \and
Jian Peng\inst{1} \and
Lei Zhang\inst{2}
}

\authorrunning{Y. Zhong et al.}
\institute{
University of Illinois at Urbana-Champaign 
\email{\{yuanyiz2,jianpeng\}@illinois.edu}
\and Microsoft
\email{\{jianfw,leizhang\}@microsoft.com}}

\fi

\maketitle

\begin{abstract}
In this paper, we propose an effective knowledge transfer framework to boost the weakly supervised object detection accuracy with the help of an external fully-annotated source dataset, whose categories may not overlap with the target domain. This setting is of great practical value due to the existence of many off-the-shelf detection datasets. To more effectively utilize the source dataset, we propose to iteratively transfer the knowledge from the source domain by a one-class universal detector and learn the target-domain detector. The box-level pseudo ground truths mined by the target-domain detector in each iteration effectively improve the one-class universal detector. Therefore, the knowledge in the source dataset is more thoroughly exploited and leveraged. Extensive experiments are conducted with Pascal VOC 2007 as the target weakly-annotated dataset and COCO/ImageNet as the source fully-annotated dataset. With the proposed solution, we achieved an mAP of $59.7\%$ detection performance on the VOC test set and an mAP of $60.2\%$ after retraining a fully supervised Faster RCNN with the mined pseudo ground truths. This is significantly better than any previously known results in related literature and sets a new state-of-the-art of weakly supervised object detection under the knowledge transfer setting. Code: \url{https://github.com/mikuhatsune/wsod_transfer}.

\end{abstract}

\section{Introduction}

Thanks to the development of powerful CNNs and novel architectures, the performance of object detectors has been dramatically improved in recent years \cite{he2016deep,girshick2015fast,ren2015faster,zhong2020anchor}. However, such successes heavily rely on supervised learning with fully annotated detection datasets which can be costly to obtain, since annotating locations and category labels of all object instances is time-consuming and sometimes prohibitively expensive. 
This issue has motivated many prior works on weakly supervised object detection (WSOD), where only image-level labels are available and normally much cheaper to obtain than box-level labels.

Existing WSOD methods \cite{song2014learning,bilen2016weakly,tang2017multiple,tang2018pcl} are mostly based on multiple instance learning (MIL), in which an image is represented as a bag of regions, e.g., generated by selective search \cite{uijlings2013selective}. The training algorithm needs to infer which instances in a bag are positive for a positive image-level class. Thus, the problem of learning a detector is converted into training an MIL classifier. 

Compared to fully supervised detectors, a large performance gap exists for weakly supervised detectors. For example, on the Pascal VOC 2007 dataset \cite{everingham2010pascal}, a fully supervised Faster RCNN can achieve an mAP of $69.9\%$~\cite{ren2015faster}, while the state-of-the-art weakly supervised detector, to the best of our knowledge, can only reach to an mAP of 53.6\% \cite{zeng2019wsod2}. 

One direction to bridge the performance gap is to utilize in a domain transfer learning setting the well-annotated external source datasets, many of which are publicly available on the web, e.g., COCO \cite{lin2014microsoft}, ImageNet \cite{imagenet_cvpr09}, Open Images \cite{kuznetsova2018open}, and Object 365 \cite{shao2019objects365}. 
Due to the existence of these off-the-shelf detection datasets, this domain transfer setting is of great practical value and has motivated many prior works, under the name transfer learning \cite{deselaers2012weakly,tang2017visual,shi2017weakly,zhang2019leveraging,kuen2019scaling,lee2018universal}, domain adaptation \cite{hoffman2014lsda,li2016weakly,chen2018lstd,hoffman2015detector,hoffman2016large}, and mixed supervised detection \cite{zhang2018mixed}. 
For example, \cite{uijlings2018revisiting} proposes to train a generic proposal generator on the source domain and an MIL classifier on the target domain in a one-step transfer manner. 
In~\cite{lee2018universal}, a universal bounding box regressor is trained on the source domain and used to refine bounding boxes for a weakly supervised detector. In~\cite{zhang2018mixed}, a domain-invariant objectness predictor is utilized to filter distracting regions before applying the MIL classifier. Other related works include \cite{deselaers2012weakly,tang2017visual,shi2017weakly,zhang2019leveraging,kuen2019scaling,li2016weakly,chen2018lstd,hoffman2016large}.

Although the domain transfer idea is very promising, it is worth noting that the top pure weakly supervised detector \cite{zeng2019wsod2} actually outperforms the best transfer-learned weakly supervised detector \cite{zhang2018mixed,lee2018universal} on VOC in the literature. 
Despite many challenges in domain transfer, one technical deficiency particularly related to object detection lies in imperfect annotations, where the source images may contain objects of the target domain categories but unannotated. In such cases, the object instances will be treated as background regions (or false negatives) in the source data, which is known as the incomplete label problem in object detection \cite{DBLP:conf/bmvc/WuBSNCD19}.
As a result, detectors trained with the source data will likely have a low recall of objects of interest in the target domain.

To address this problem, we propose to transfer progressively so that the knowledge can be extracted more thoroughly by taking into account the target domain. 
Specifically, we iterate between extracting knowledge by a one-class universal detector (OCUD) and learning a target domain object detector through MIL.
The target domain detector is used to mine the pseudo ground truth annotations in both the source and target datasets to refine the OCUD. 
Compared with existing works, the key novelty is to extract knowledge in multi-steps rather than one-step. Technically, by adding pseudo ground truths in the source data, we effectively alleviate the problem of false negatives as aforementioned.
By adding pseudo ground truths in and including the target dataset in fine-tuning, the refined OCUD is more adapted to the target domain data distribution.
Empirically, we observe significant gains, e.g., from $54.93\%$ mAP with one-step transfer to $59.71\%$ with multi-step transfer (5 refinements) on Pascal VOC 2007 test data by leveraging COCO-60 as source (removing the VOC 20 categories). 
By retraining a fully supervised Faster RCNN with the mined pseudo ground truths, we can achieve 60.24\% mAP, which again surpasses the pure WSOD method~\cite{zeng2019wsod2} remarkably and sets a new state of the art under the transfer setting. Finally, as a reference, the detection performance also surpasses the original fully supervised Faster RCNN with the ZF net backbone (59.9\% mAP) \cite{ren2015faster}.

\section{Related Work}

\paranoskip{Weakly Supervised Object Detection (WSOD).} WSOD is extensively studied in the literature \cite{song2014learning,bilen2016weakly,tang2017multiple,tang2018pcl}. 
The problem is often formulated as an image classification with multi-instance learning. 
Typically, candidate bounding boxes are first generated by independent proposal methods such as Edge Boxes \cite{zitnick2014edge} and Selective Search \cite{uijlings2013selective}. Then the proposals on one image are treated as a bag with the image labels as bag-level labels.
WSDDN \cite{bilen2016weakly} utilizes a two-stream architecture that separates the detection and classification scores, which are then aggregated through softmax pooling to predict the image labels. 
OICR \cite{tang2017multiple} and the subsequent PCL \cite{tang2018pcl} transform the image-level labels into instance-level labels by multiple online classifier refinement steps.
Class activation maps can also be used to localize objects \cite{zhu2017soft}. WSOD2 \cite{zeng2019wsod2} exploits the bottom-up and top-down objectness to improve performance. Among existing works, pseudo ground truth mining is heavily used as a tool for iterative refinement \cite{tang2018weakly,tang2018pcl,zhang2018w2f}.

Classifier refinement methods such as OICR \cite{tang2017multiple} and PCL \cite{tang2018pcl} are related in that they conduct refinement steps. Our method is similar to them when restricted to operating on the target data only. However, there are several notable differences. We study the WSOD-with-transfer rather than the pure WSOD setting. Our pseudo ground truth mining is conducted on both the source and target data. We refine both the classifier and the box proposals by retraining the OCUD rather than the instance classifier only.

\para{WSOD with Knowledge Transfer.}
One way to improve the accuracy is to utilize a source dataset and transfer the knowledge to the target domain through semi-supervised or transfer learning. 
Visual or semantic information in the category labels or images is often exploited to help solve the problem.
For example, the word embeddings of category texts are employed in \cite{tang2017visual,bansal2018zero} to represent class semantic relationships.
The appearance model learned on the source classes are transferred to the target classes in \cite{shi2017weakly,rochan2015weakly,li2016weakly}.
Many methods leverage weight prediction to effectively turn a novel category classifier into a detector \cite{kuen2019scaling,hoffman2014lsda,tang2017visual}.
For example, LSDA \cite{hoffman2014lsda} and \cite{tang2017visual} transfer the classifier-to-detector weight differences. 
Recent works \cite{deselaers2012weakly,uijlings2018revisiting,lee2018universal,zhang2018mixed} share with us in spirit learning general object knowledge from the source data. The knowledge can either be the objectness predictor \cite{deselaers2012weakly,zhang2018mixed}, the object proposals \cite{uijlings2018revisiting} or the universal bounding box regressor \cite{lee2018universal}. 
In particular, \cite{uijlings2018revisiting} also trains a universal detector (in their case, SSD \cite{liu2016ssd}) on the source dataset, and uses the detection results from this detector as proposals during MIL on the target dataset. The process can be seen as a special case of our algorithm with a single-step transfer and a different instantiation of network and MIL method.
Comparatively, we differentiate our approach from them by progressively exploiting the knowledge in the source dataset in a multi-step way, such that the accuracy can improve gradually. 
Empirically, we observed non-trivial performance gain with progressive knowledge transfer. 

\section{Proposed Approach}

Given source dataset $\cS$ with bounding box annotations and target dataset $\cT$ with only image-level labels, the goal is to train an object detector for object categories in $\cT$.
The categories of $\cS$ and $\cT$ can be non-overlapping, which differentiates our setting from a typical semi-supervised setting. 

\begin{figure*}[t]
\begin{center}
\includegraphics[width=\linewidth]{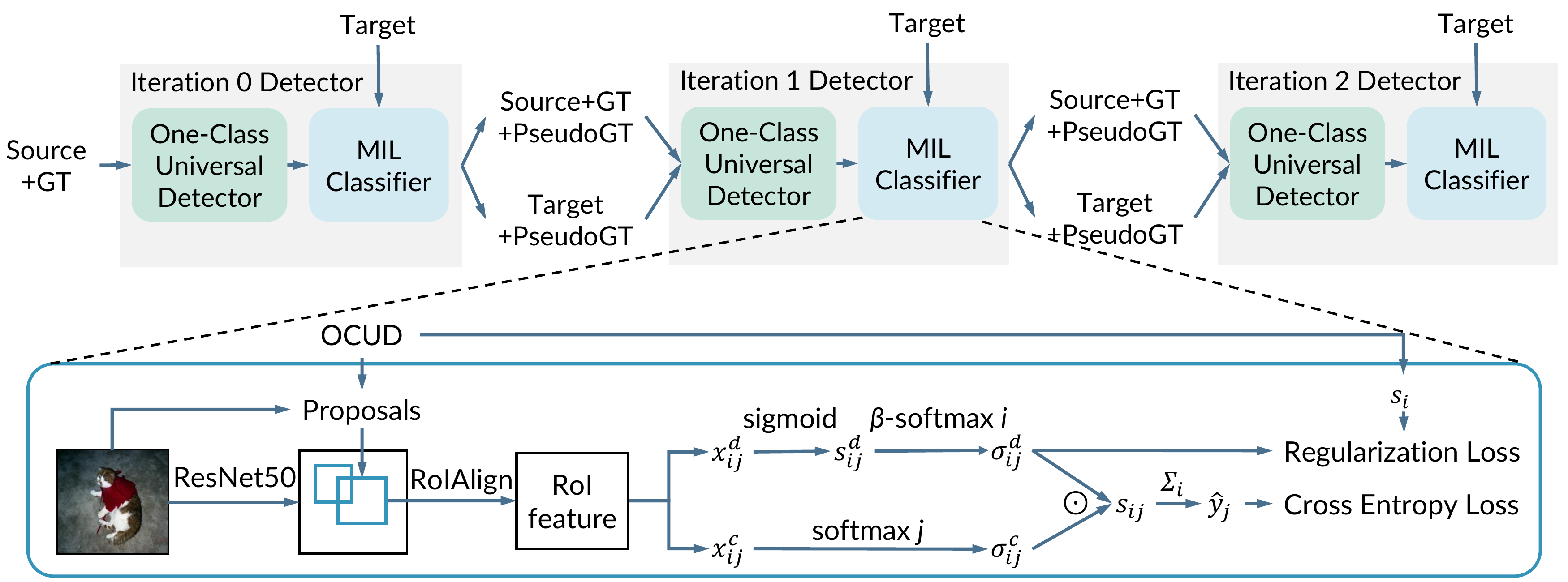}
\end{center}
\caption{An illustration of the proposed progressive knowledge transfer framework.
One-class universal detector (OCUD) is initially trained with fully annotated source data and iteratively refined on source and target data with pseudo ground truths (GT). OCUD acts as the proposal generator during the subsequent training of target domain MIL classifiers. OCUD and MIL classifier together form the target domain detector.}
\label{fig:pipeline}
\end{figure*}

The proposed training framework and workflow are outlined in Fig.~\ref{fig:pipeline} and Alg.~\ref{algo:main}. 
The basic flow is to first train a target domain detector as a seed based on the existing labels, and then mine the pseudo ground truth boxes, which are then used to refine the detector. 
The process is repeated to improve the target domain detector gradually
since more target domain boxes can be found in both $\cS$ and $\cT$ through the process.
The architecture design of the detector is versatile. 
Here we present a simple solution consisting of a one-class universal detector (OCUD) and a MIL classifier. 

\begin{algorithm}[t]
\label{algo:main}
\small
\KwIn{ Max number of refinements $N$, source dataset $\cS$, target dataset $\cT$; }
Train the one-class universal detector (OCUD) on the source dataset $\cS$\;
Train the MIL classifier based on the OCUD and the target dataset $\cT$\;
\For{$K = 1,2,\ldots N$} {
    Mine pseudo ground truths in $\cS$ and $\cT$ with OCUD and the MIL classifier\;
    Refine the OCUD with the mined boxes and original source annotations\;
    Refine the MIL classifier based on the OCUD and the target dataset $\cT$\;
}
\Return The OCUD and MIL classifier as the target domain detector\;
\caption{WSOD with Progressive Knowledge Transfer.}
\end{algorithm}

\subsection{One-Class Universal Detector (OCUD)}
The one-class universal detector, which we refer to as OCUD for convenience, treats all categories as a single generic category. While we employ Faster RCNN \cite{ren2015faster} with ResNet50 \cite{he2016deep} backbone, any modern object detector can be used.

Initially, the OCUD is trained on source data only, which is similar to \cite{uijlings2018revisiting}. 
Although the categories can be non-overlapping between the source domain and the target domain, the objects may be visually similar to some extent, which gives the detector certain capability to detect the target domain objects. For example, a detector trained on \textit{cat} might be able to detect \textit{dog}.

\subsection{MIL Classifier}
\label{sec:method.2}
With the OCUD, we extract multiple proposals in the target dataset image and perform multiple instance learning (MIL) with the proposals. 
Our MIL classifier is based on WSDDN \cite{bilen2016weakly}, but adapted to incorporate knowledge from the OCUD.

\def\dij{^\text{d}_{ij}}
\def\cij{^\text{c}_{ij}}

The MIL classifier has a two-stage Faster-RCNN-like architecture sketched in Fig.~\ref{fig:pipeline}.
Assume that the OCUD gives $R$ proposals in a target dataset image: $\cbr{ b_i, s_i }_{i=1}^R$. 
We run RoIAlign \cite{he2017mask} to extract a feature map for each proposal, and feed the feature into two branches as in \cite{bilen2016weakly}: the detection branch and the classification branch. 
Each branch consists of 2 linear layers with ReLU. 
The last linear layer's output has the same dimension as the number of target domain categories. 
Let $x\dij \in \bR$ and $x\cij \in \bR$ be the output for the $i$-th proposal and the $j$-th category from the detection branch and the classification branch, respectively. 
The predicted score $s_{ij}$ is calculated as follows, 
\begin{equation}
\begin{split}
\label{eq:softmax_score}
s\dij = \sigmoid(x\dij)&,
\quad
\sigma\dij = \softmax_i( \beta s\dij ),
\\
\sigma\cij = \softmax_j( x\cij )&,
\quad
s_{ij} = \sigma\dij \sigma\cij
.
\end{split}
\end{equation}

The $\softmax$ is computed along the $i$ and $j$ dimensions respectively. Different from \cite{bilen2016weakly}, we squash the detection scores $s\dij$ to $(0,1)$ by sigmoid. 
This has two benefits: (1) It allows multiple proposals to belong to the same category: $s\dij$ represents how likely each proposal individually belongs to category $j$, and $\sigma\dij$ is a normalization; (2) It makes it easier to enforce the objectness regularization as we shall see below. 
To make $\sigma\dij$ amenable to train, we introduce a scaling factor $\beta$ to adjust the input range from $(0, 1)$ to $(0, \beta)$.
With a larger $\beta$, the range of the scaled softmax is wider, and the value is easier to be spiked. 


Let $\{y_j\}_{j=1}^C \in \{0,1\}^C$ be the image-level label, and $C$ be the number of categories. Given the scores of all proposals, an image-level classification prediction $\hat{y}_j$ is calculated and used in an image-level binary classification loss $L_{\text{wsddn}}$,
\begin{equation}
    \hat{y}_j = \sum_{i=1}^R s_{ij}, \quad
    L_{\text{wsddn}} = - \frac1C \sum_{j=1}^C y_j \log \hat{y}_j + (1-y_j) \log (1-\hat{y}_j) .
    \label{eq:wsddn}
\end{equation}

To further exploit the knowledge present in OCUD, we introduce the following $L_2$ regularization loss on the detection branch scores $s\dij$. 
The intuition behind is that the objectness score $s_i$ predicted by the OCUD could guide MIL by promoting the object candidates' confidence. It should match the target domain detector's objectness of region $i$ defined as the max over classes. The overall training loss for each image is the weighted sum with coefficient $\lambda$ as in Eq.~\ref{eq:loss}.
\begin{equation}
    L_{\text{guide}} = \frac1R \sum_{i=1}^R \big( \max_{1\le j\le C} s\dij - s_i \big)^2 .
    \label{eq:guide}
\end{equation}
\begin{equation}
    \cL = L_{\text{wsddn}} + \lambda L_{\text{guide}}.
    \label{eq:loss}
\end{equation} 

During inference, the final detection score is the linear interpolation of $s_i$ from the OCUD and $s_{ij}$ from the MIL classifier. 
This scheme is shown to be robust  \cite{uijlings2018revisiting}. 
Specifically, with a coefficient $\eta \in [0,1]$, we compute the final score by Eq.~\ref{eq:final_score}. The model trusts the MIL classifier more with a larger $\eta$.
\begin{equation}
    s^{\text{final}}_{ij} = \eta s_{ij} + (1-\eta) s_i .
    \label{eq:final_score}
\end{equation}

\subsection{Pseudo Ground Truth Mining}

\def\vvanno{\mathrm{\textit{anno}}}
\def\vvscore{\mathrm{\textit{score}}}
\def\vvscores{\mathrm{\textit{scores}}}
\def\vvbox{\mathrm{\textit{box}}}
\def\vvov{\mathrm{\textit{overlaps}}}

\begin{algorithm}[tb]
\label{algo:pseudo}
\small
\KwIn{Detector $D_{\cT}$, source $\cS$, target $\cT$, score threshold $\tau$, overlap threshold $o$}
$\cS^+ \leftarrow \emptyset$, $\cT^+ \leftarrow \emptyset$\;
\For{\textup{(image $I$, boxes $B$) in $\cS$}} {
    predictions $P \leftarrow D_{\cT}(I)$;
    annotations $\vvanno \leftarrow B$\;
    \For{\textup{predicted box $p$ in $P$}} {
        \If{\textup{$p.\vvscore > \tau$}} {
            $\vvov \leftarrow \operatorname{overlap}(p.\vvbox, B) / \operatorname{area}(p.\vvbox)$\;
            \lIf{$\max \vvov < o$} {
                add $p$ to $\vvanno$
            }
        }
    }
    add $(I,\vvanno)$ to $\cS^+$\;
}
\For{\textup{(image $I$, image label $Y$) in $\cT$}} {
    predictions $P \leftarrow D_{\cT}(I)$;
    annotations $\vvanno \leftarrow \emptyset$\;
    \For{\textup{category $y$ in $Y$}} {
        find subset predictions $P_y \leftarrow \{p\in P: p.\mathrm{\textit{category}} = y \}$\;
        \For{\textup{box $p$ in $P_y$}} {
            \lIf{\textup{$p.\vvscore > \tau$ or $p.\vvscore = \max P_y.\vvscores$}} {
                add $p$ to $\vvanno$
            }
        }
    }
    add $(I,\vvanno)$ to $\cT^+$\;
}
\Return $\cS^+$, $\cT^+$\;
\caption{Pseudo Ground Truth Mining.}
\end{algorithm}

Given the OCUD and the MIL classifier, we mine the pseudo ground truth on both the source and the target dataset based on the latest target domain detector (OCUD + MIL classifier).
Following \cite{tang2017multiple,zeng2019wsod2,jie2017deep}, we adopt the simple heuristic to pick the most confident predictions, as summarized in Alg.~\ref{algo:pseudo}.

In the source dataset, the predictions with high confidence (thresholded by $\tau$) and low overlap ratio (thresholded by $o$) with the nearest ground truth bounding box are taken as a pseudo ground truth. 
Here we use the intersection over the predicted bounding box area as the overlap ratio, to conservatively mine the box in the source data and avoid mining object parts. 
Empirically, this simple scheme is effective to locate target domain objects in the source dataset. 

In the target dataset, the image-level labels are used to filter the predictions in addition to the confidence scores. For each positive class, we select as pseudo ground truth the top one box and any detection result with a confidence score higher than the threshold $\tau$. In this way, any misclassified bounding box is filtered out, and each positive class is guaranteed to have at least one box.

\subsection{Refinement of OCUD and MIL Classifier}

Pseudo ground truth augmented source and target datasets are used to refine the OCUD. The fine-tuning is the same as the initial OCUD training, except that the two domain images are now mixed together, and the model is initialized from the last OCUD. More advanced techniques can be leveraged, e.g., assigning different weights for the pseudo ground truth in the target dataset, the source dataset, and the original source annotations. We leave it as future work. 

In the experiments, we find this simple refinement approach is effective. 
Through the last target domain detector, the mined pseudo ground truth boxes are better aligned towards the target domain categories. 
In the target dataset, the objects could be correctly localized, and the boxes become the pseudo ground truths to improve the OCUD.
In the source dataset, the pseudo ground truth can improve the recall rate, especially when the image content contains the target category (not in the source domain category).
Without refinement, those regions will be treated as the background, which is detrimental.

With the improved OCUD, the MIL classifier is also fine-tuned by the improved object proposals detected by the OCUD. 
Before the refinements, the OCUD contains little information on the target domain categories, and the proposals are generated by solely relying on the similarity of the categories across domains (e.g., being able to detect \textit{horse} might help detect \textit{sheep}). 
Afterwards, the OCUD is improved to incorporate more information about the target domain, and the proposals will also likely be aligned to improve the MIL classifier. 

\section{Experiments}
\label{sec:expr}

\subsection{Experiment Settings}
\paranoskip{Target Dataset.}
Following \cite{lee2018universal,zhang2018mixed}, we use Pascal VOC 2007 dataset \cite{everingham2010pascal} as the target dataset, which has 2501 training images with 6301 box-level annotations, 2510 validation images with 6,307 annotations and 4,952 testing images with 12,032 annotations.
As in \cite{lee2018universal,zhang2018mixed,bilen2016weakly,tang2018pcl,zeng2019wsod2}, we combine the training and validation sets into one trainval set for training, and evaluate the accuracy on the test set. 
The bounding boxes are removed in the trainval set, and only the image-level labels are kept for the weakly supervised training. 
There are 20 categories. 

\para{Source Dataset.}
Similar to \cite{lee2018universal}, we use COCO \cite{lin2014microsoft} 2017 detection dataset as the source dataset, which contains 118,287 training images with 860,001 box-level annotations and 5,000 validation images with 36,781 annotations. 
The number of categories is 80, and all the 20 categories of VOC are covered.
As in~\cite{lee2018universal}, we remove all the images that have overlapped categories with VOC, resulting in a train set of 21,987 images with 70,549 annotations, and a validation set of 921 images with 2,914 annotations. 
The resulting train and validation sets are merged as the source dataset, which we denote as COCO-60. 
We aim to transfer the knowledge through the one-class universal detector from the COCO-60 dataset to the weakly labeled VOC dataset with no overlapping classes.

Another source dataset we investigate is ILSVRC 2013 detection dataset, which contains 395,909 train images (345,854 box annotations) and 20,121 validation images (55,502 box annotations) of 200 classes. After removing images of the 21\footnote{ILSVRC has 2 classes \textit{water bottle} and \textit{wine bottle} while COCO and VOC have \textit{bottle}.} categories overlapping with VOC, we arrive at 143,095 train images and 6,229 val images of 179 classes.
The train and validation images are combined as the source dataset, denoted as ILSVRC-179 in the ablation study. Without an explicit description, we use COCO-60 as the source dataset. 

\para{Evaluation Metrics.}
We adopt two evaluation metrics frequently used in weakly supervised detection literature, namely mean average precision (mAP) and Correct Localization (CorLoc).
Average precision (AP) is the area under the precision/recall curve
for each category, and mAP averages the APs of all categories. 
CorLoc \cite{deselaers2012weakly} measures the localization accuracy on the training dataset. It is defined as the percentage of images of a certain class that the top one prediction of the algorithm correctly localizes one object. A prediction is correct if the intersection-over-union (IoU) with ground truth is larger than 0.5.

\para{Network.}
We use the Faster RCNN as the one-class universal detector (OCUD), where the RPN network is based on the first 4 conv stages of ResNet, and the RoI CNN is based on the 5th conv stage. 
It is worth noting that any detector can be used here. 
Up to 100 detected boxes are fed from the OCUD to the MIL classifier. 
ResNet50 is used as the backbone for both OCUD and the MIL classifier.
Our implementation is based on maskrcnn-benchmark \cite{massa2018maskrcnn}.

\para{Training and Inference. }
The training is distributed over 4 GPUs, with a batch size of 8 images. 
The OCUD is initialized with the ImageNet pre-trained model and trained with 17,500 steps. Afterwards, the OCUD is fine-tuned with 5000 steps in the refinements. 
The MIL classifier is trained for 5000 steps initially and then fine-tuned similarly for 2000 steps in each following refinement. 
The base learning rate is set to 0.008 for all experiments and all models and is reduced by 0.1 after finishing roughly 70\% of the training progress. 

It is worth noting that the overhead training time of $K$ refinements is less than $K$ times the usual training time, due to the shortened training schedule for the refinements. For example, in the COCO-60-to-VOC experiments, the initial OCUD and MIL training cost 190min and 23min, but the OCUD and MIL refinements only took 50min and 8min. 
The testing time of the final distilled detector is similar to the usual detector. The details are in the supplementary.

Without explicit description, the parameter $\beta$ in Eq.~\ref{eq:softmax_score} is 5, the $\lambda$ in Eq.~\ref{eq:loss} is 0.2, the $\eta$ in Eq.~\ref{eq:final_score} is 0.5, and the number of refinements is 5.
In the phase of pseudo ground truth mining, the confidence threshold $\tau$ is 0.8, and the IoU threshold $o$ is 0.1.
We also studied the sensitivity of these parameters.

The training images are resized to have a short edge of 640 pixels. During testing, we study both the single-scale no-augmentation configuration, and the multi-scale (two: 320, 640 pixels) setting with horizontal flipping as adopted in prior work \cite{zeng2019wsod2,zhang2018mixed,tang2018pcl}. 
The non-maximum suppression IoU is 0.4 during testing.

\subsection{Comparison with SOTA}

\begin{table}[t]
    \caption{mAP performance on VOC 2007 test set. `Ours' are trained with COCO-60 as source. Superscript `+' indicates multi-scale testing. `Distill' means to re-train a Faster RCNN based on the mined boxes. `Ens' indicates ensemble methods.}
    \scalebox{0.68}{
    \setlength{\tabcolsep}{1pt}
    \hskip-4pt
    \begin{tabular}{@{} l c*{20} c}
        \toprule
        Method & aero & bike & bird & boat & bottl & bus & car & cat & chair & cow & table & dog & horse & mbik & pers. & plant & sheep & sofa & train & tv & mAP \\
        \midrule
        \textbf{Pure WSOD:} \\
        WSDDN-Ens \cite{bilen2016weakly} & 46.4 & 58.3 & 35.5 & 25.9 & 14.0 & 66.7 & 53.0 & 39.2 & 8.9 & 41.8 & 26.6 & 38.6 & 44.7 & 59.0 & 10.8 & 17.3 & 40.7 & 49.6 & 56.9 & 50.8 & 39.3 \\
        OICR-Ens+FR \cite{tang2017multiple} & 65.5 & 67.2 & 47.2 & 21.6 & 22.1 & 68.0 & 68.5 & 35.9 & 5.7 & 63.1 & 49.5 & 30.3 & 64.7 & 66.1 & 13.0 & 25.6 & 50.0 & 57.1 & 60.2 & 59.0 & 47.0 \\
        PCL-Ens+FR \cite{tang2018pcl} & 63.2 & 69.9 & 47.9 & 22.6 & 27.3 & 71.0 & 69.1 & 49.6 & 12.0 & 60.1 & 51.5 & 37.3 & 63.3 & 63.9 & 15.8 & 23.6 & 48.8 & 55.3 & 61.2 & 62.1 & 48.8 \\
        WSOD2\textsuperscript{+} \cite{zeng2019wsod2} & 65.1 & 64.8 & 57.2 & 39.2 & 24.3 & 69.8 & 66.2 & 61.0 & 29.8 & 64.6 & 42.5 & 60.1 & 71.2 & 70.7 & 21.9 & 28.1 & 58.6 & 59.7 & 52.2 & 64.8 & 53.6 \\
        \textbf{With transfer:} \\
        MSD-Ens\textsuperscript{+} \cite{zhang2018mixed} & 70.5 & 69.2 & 53.3 & 43.7 & 25.4 & 68.9 & 68.7 & 56.9 & 18.4 & 64.2 & 15.3 & 72.0 & 74.4 & 65.2 & 15.4 & 25.1 & 53.6 & 54.4 & 45.6 & 61.4 & 51.08 \\
        OICR+UBBR \cite{lee2018universal} & 59.7 & 44.8 & 54.0 & 36.1 & 29.3 & 72.1 & 67.4 & 70.7 & 23.5 & 63.8 & 31.5 & 61.5 & 63.7 & 61.9 & 37.9 & 15.4 & 55.1 & 57.4 & 69.9 & 63.6 & 52.0 \\
        \midrule
        \textbf{Ours:} \\
        Ours(single scale) & 64.4 & 45.0 & 62.1 & 42.8 & 42.4 & 73.1 & 73.2 & 76.0 & 28.2 & 78.6 & 28.5 & 75.1 & 74.6 & 67.7 & 57.5 & 11.6 & 65.6 & 55.4 & 72.2 & 61.3 & 57.77 \\
        Ours\textsuperscript{+} & 64.8 & 50.7 & 65.5 & 45.3 & 46.4 & 75.7 & 74.0 & 80.1 & 31.3 & 77.0 & 26.2 & 79.3 & 74.8 & 66.5 & 57.9 & 11.5 & 68.2 & 59.0 & 74.7 & 65.5 & 59.71 \\
        Ours(distill,vgg16)\textsuperscript{+} & 62.6 & 56.1 & 64.5 & 40.9 & 44.5 & 74.4 & 76.8 & 80.5 & 30.6 & 75.4 & 25.5 & 80.9 & 73.4 & 71.0 & 59.1 & 16.7 & 64.1 & 59.5 & 72.4 & 68.0 & 59.84 \\
        Ours(distill)\textsuperscript{+} & 65.5 & 57.7 & 65.1 & 41.3 & 43.0 & 73.6 & 75.7 & 80.4 & 33.4 & 72.2 & 33.8 & 81.3 & 79.6 & 63.0 & 59.4 & 10.9 & 65.1 & 64.2 & 72.7 & 67.2 & 60.24 \\
        \midrule
        \textbf{Upper bounds:} \\
        Fully Supervised  & 75.9 & 83.0 & 74.4 & 60.8 & 56.5 & 79.0 & 83.8 & 83.6 & 54.9 & 81.6 & 66.8 & 85.3 & 84.3 & 77.4 & 82.6 & 47.3 & 74.0 & 72.2 & 78.0 & 74.8 & 73.82 \\
        Ideal OCUD & 70.0 & 72.4 & 72.6 & 51.7 & 57.5 & 76.1 & 80.7 & 86.8 & 45.8 & 81.3 & 50.6 & 81.6 & 78.4 & 72.5 & 74.4 & 45.4 & 70.1 & 61.5 & 76.0 & 72.9 & 68.92 \\
        \bottomrule
    \end{tabular}
    }
    \label{tbl:coco_voc_map}
\end{table}

\begin{table}[t]
    \caption{CorLoc performance on VOC 2007 trainval set. `Ours' are trained with COCO-60 as source. Superscript `+' indicates multi-scale testing. `Distill' means to re-train a Faster RCNN based on the mined boxes. `Ens' indicates ensemble methods.}
    \scalebox{0.68}{
    \setlength{\tabcolsep}{1pt}
    \hskip-4pt
    \begin{tabular}{@{} l c*{20} c}
        \toprule
        Method & aero & bike & bird & boat & bottl & bus & car & cat & chair & cow & table & dog & horse & mbik & pers. & plant & sheep & sofa & train & tv & Cor.\\
        \midrule
        \textbf{Pure WSOD:} \\
        WSDDN-Ens \cite{bilen2016weakly} & 68.9 & 68.7 & 65.2 & 42.5 & 40.6 & 72.6 & 75.2 & 53.7 & 29.7 & 68.1 & 33.5 & 45.6 & 65.9 & 86.1 & 27.5 & 44.9 & 76.0 & 62.4 & 66.3 & 66.8 & 58.0 \\
        OICR-Ens+FR \cite{tang2017multiple} & 85.8 & 82.7 & 62.8 & 45.2 & 43.5 & 84.8 & 87.0 & 46.8 & 15.7 & 82.2 & 51.0 & 45.6 & 83.7 & 91.2 & 22.2 & 59.7 & 75.3 & 65.1 & 76.8 & 78.1 & 64.3 \\
        PCL-Ens+FR \cite{tang2018pcl} & 83.8 & 85.1 & 65.5 & 43.1 & 50.8 & 83.2 & 85.3 & 59.3 & 28.5 & 82.2 & 57.4 & 50.7 & 85.0 & 92.0 & 27.9 & 54.2 & 72.2 & 65.9 & 77.6 & 82.1 & 66.6 \\
        WSOD2\textsuperscript{+} \cite{zeng2019wsod2} & 87.1 & 80.0 & 74.8 & 60.1 & 36.6 & 79.2 & 83.8 & 70.6 & 43.5 & 88.4 & 46.0 & 74.7 & 87.4 & 90.8 & 44.2 & 52.4 & 81.4 & 61.8 & 67.7 & 79.9 & 69.5 \\
        \textbf{With transfer:} \\
        WSLAT-Ens \cite{rochan2015weakly} & 78.6 & 63.4 & 66.4 & 56.4 & 19.7 & 82.3 & 74.8 & 69.1 & 22.5 & 72.3 & 31.0 & 63.0 & 74.9 & 78.4 & 48.6 & 29.4 & 64.6 & 36.2 & 75.9 & 69.5 & 58.8 \\
        MSD-Ens\textsuperscript{+} \cite{zhang2018mixed} & 89.2 & 75.7 & 75.1 & 66.5 & 58.8 & 78.2 & 88.9 & 66.9 & 28.2 & 86.3 & 29.7 & 83.5 & 83.3 & 92.8 & 23.7 & 40.3 & 85.6 & 48.9 & 70.3 & 68.1 & 66.8 \\
        OICR+UBBR \cite{lee2018universal} & 47.9 & 18.9 & 63.1 & 39.7 & 10.2 & 62.3 & 69.3 & 61.0 & 27.0 & 79.0 & 24.5 & 67.9 & 79.1 & 49.7 & 28.6 & 12.8 & 79.4 & 40.6 & 61.6 & 28.4 & 47.6 \\
        \midrule
        \textbf{Ours:} \\
        Ours(single scale) & 86.7 & 62.4 & 87.1 & 70.2 & 66.4 & 85.3 & 87.6 & 88.1 & 42.3 & 94.5 & 32.3 & 87.7 & 91.2 & 88.8 & 71.2 & 20.5 & 93.8 & 51.6 & 87.5 & 76.7 & 73.6 \\
        Ours\textsuperscript{+} & 87.5 & 64.7 & 87.4 & 69.7 & 67.9 & 86.3 & 88.8 & 88.1 & 44.4 & 93.8 & 31.9 & 89.1 & 92.9 & 86.3 & 71.5 & 22.7 & 94.8 & 56.5 & 88.2 & 76.3 & 74.4 \\
        Ours(distill,vgg16)\textsuperscript{+} & 87.9 & 66.7 & 87.7 & 67.6 & 70.2 & 85.8 & 89.9 & 89.2 & 47.9 & 94.5 & 30.8 & 91.6 & 91.8 & 87.6 & 72.2 & 23.8  & 91.8 & 67.2 & 88.6 & 81.7 & 75.7 \\
        Ours(distill)\textsuperscript{+} & 85.8 & 67.5 & 87.1 & 68.6 & 68.3 & 85.8 & 90.4 & 88.7 & 43.5 & 95.2 & 31.6 & 90.9 & 94.2 & 88.8 & 72.4 & 23.8 & 88.7 & 66.1 & 89.7 & 76.7 & 75.2 \\
        \midrule
        \textbf{Upper bounds:} \\
        Fully Supervised & 99.6 & 96.1 & 99.1 & 95.7 & 91.6 & 94.9 & 94.7 & 98.3 & 78.7 & 98.6 & 85.6 & 98.4 & 98.3 & 98.8 & 96.6 & 90.1 & 99.0 & 80.1 & 99.6 & 93.2 & 94.3 \\
        Ideal OCUD & 97.5 & 85.1 & 96.7 & 83.5 & 84.4 & 91.9 & 92.5 & 94.5 & 65.4 & 95.2 & 70.0 & 94.2 & 94.6 & 91.6 & 90.6 & 81.3 & 96.9 & 61.3 & 96.6 & 88.2 & 87.6 \\
        \bottomrule
    \end{tabular}
    }
    \label{tbl:coco_voc_corloc}
\end{table}

Table~\ref{tbl:coco_voc_map} and Table~\ref{tbl:coco_voc_corloc} compare our approach with previous state-of-the-art approaches in terms of mAP and CorLoc, respectively.  
We compare to pure WSOD methods: (1) WSDDN-Ens \cite{bilen2016weakly}, the ensemble of 3 Weakly Supervised Detection Networks. Our two branch MIL is modified from WSDDN. (2) OICR-Ens+FR \cite{tang2017multiple}, a Fast RCNN \cite{girshick2015fast} retrained from a VGG ensemble of the Online Instance Classifier Refinement models. (3) PCL-Ens+FR \cite{tang2018pcl}, an improvement over OICR \cite{tang2017multiple} which leverages proposal clusters to refine classifiers. (4) WSOD2\textsuperscript{+} \cite{zeng2019wsod2}, one of the best-performing WSODs on VOC which combines bottom-up and top-down objectness cues.
We also compare with two WSOD-with-transfer methods: (1) MSD-Ens\textsuperscript{+} \cite{zhang2018mixed} which transfers the objectness learned from source, (2) OICR+UBBR \cite{lee2018universal} which learns a universal box regressor on source data.

From the tables, we have the following observations.
\begin{enumerate}
    \item Our approach without multi-scale testing and model retraining (distilling) achieves significantly higher accuracy than any pure WSOD.
    In terms of mAP, the gain is more than 4 points from 53.6\% \cite{zeng2019wsod2} to 57.77\% (ours). For CorLoc, it is from 59.5\% \cite{zeng2019wsod2} to 73.6\%. This demonstrates the superior advantage of leveraging existing detection source dataset to help the novel or unseen weakly supervised training task.
    
    \item Compared with the approaches using external data, our approach performs consistently higher than the top related approach both in mAP and CorLoc. 
    The best previous mAP is 52.0\% \cite{lee2018universal}, and the best CorLoc is 66.8\% \cite{zhang2018mixed}. Both numbers are behind the best pure WSOD approach, and the reason might be the insufficient utilization of the external data. 
    Instead, we utilize the external data more thoroughly with multiple progressive refinements, which significantly boosts the final accuracy.
    
    \item For our approach, the multi-scale testing gives around 2 points' gain in mAP and 1 point's gain in CorLoc. 
    
    \item Similar to \cite{li2016weakly,jie2017deep,tang2018pcl}, we retrain a Faster RCNN detector on the VOC trainval images with the pseudo box annotations from our OCUD and MIL classifier. 
    With VGG16 as the backbone, the accuracy (shown with \text{distill}) is $59.84\%$ in mAP and $75.7\%$ in CorLoc. With a more powerful backbone of ResNet50, mAP is $60.24\%$ and CorLoc is $75.2\%$.
    Though the backbones are notably different, we observe the accuracy does not change accordingly, and the bottleneck may still be the quality of the mined pseudo ground truth. 
    With $60.24\%$ mAP, our approach surpasses the Faster RCNN fully supervised detector with the ZF network backbone ($59.9\%$ mAP) \cite{ren2015faster}.
    
    \item Two numbers are reported as upper bounds in Table~\ref{tbl:coco_voc_map}. The first one is a fully supervised Faster RCNN (ResNet50) based on the true box annotations, which achieves 73.82\% mAP.
    The other upper bound is estimated based on our training pipeline but with the fully annotated VOC as the source dataset.
    That is, the true ground truth bounding boxes of VOC trainval are used to train the OCUD, which yields 68.92\% mAP. 
    Thus, the gap from $60.24\%$ (our best result) to $68.92\%$ mAP may mainly come from data disparity between the source and the target, signifying room for further improvement.
    Investigating a more advanced pseudo ground truth mining approach and resorting to more source data could help close the gap in the future. 
\end{enumerate}

\begin{figure}[t]
    \centering
    \small
    \begin{tabular}{@{}c@{}c@{}c}
        \includegraphics[width=0.33\linewidth]{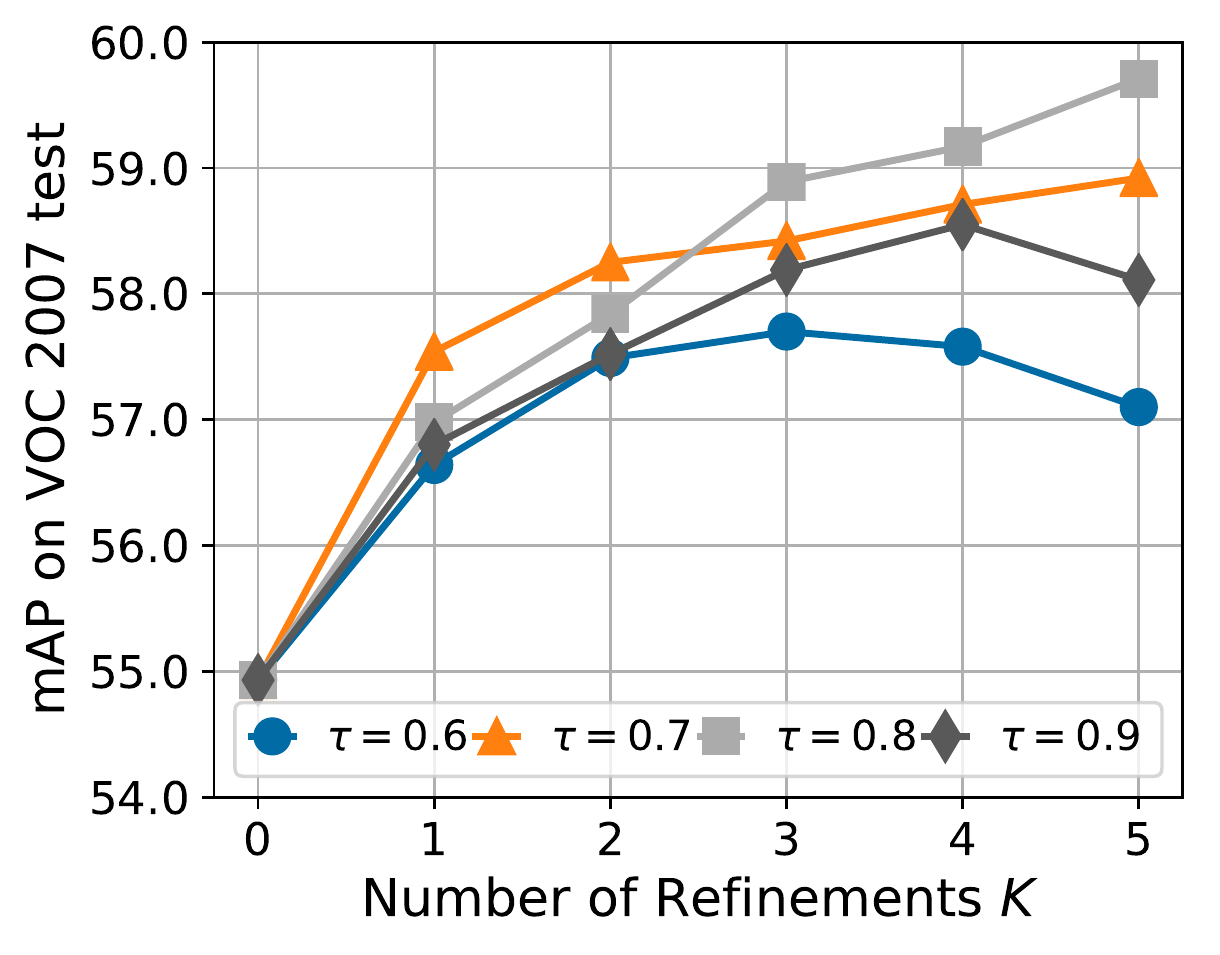} &
        \includegraphics[width=0.33\linewidth]{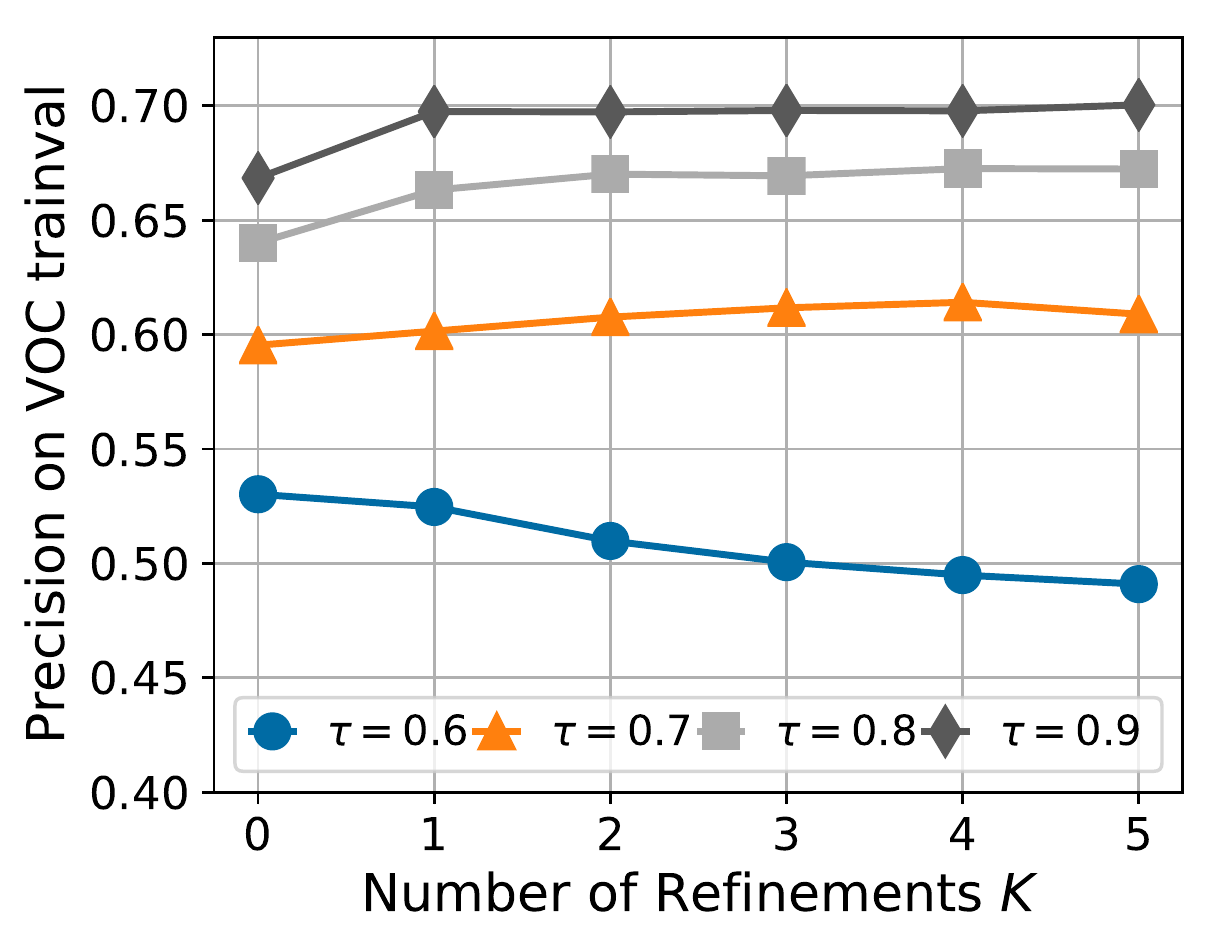} &
        \includegraphics[width=0.33\linewidth]{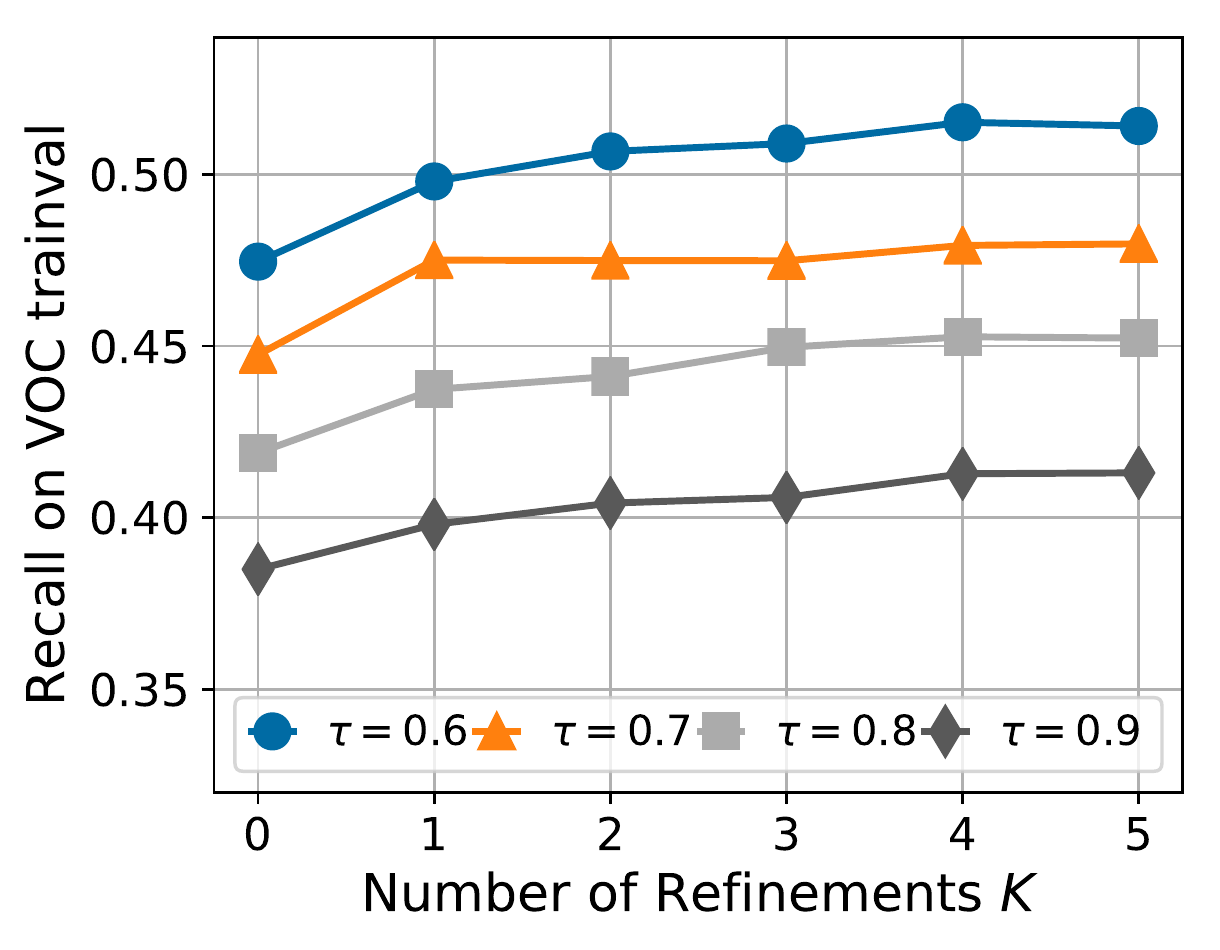}
        \\
        (a) mAP & (b) Precision & (c) Recall
    \end{tabular}
    \caption{Accuracy with different pseudo ground truth mining thresholds. 
    (a) mAP on VOC test; (b)/(c): precision/recall of the mined pseudo ground truth on VOC trainval.}
    \label{fig:tau_prec_recall}
\end{figure}

\subsection{Ablation Study}

\paranoskip{$\tau$ and $K$.}
Fig~\ref{fig:tau_prec_recall}(a) shows the mAP with multi-scale testing under different thresholds of $\tau$ (0.6,0.7,0.8,0.9) and the number of refinements $K$. 
Fig~\ref{fig:tau_prec_recall}(b) and (c) shows the corresponding precision and recall of the pseudo ground truth on the target dataset. 
The threshold $\tau$ is used in the pseudo ground truth mining in Alg.~\ref{algo:pseudo}. 
From (b) and (c), we can see a higher threshold leads to higher precision but lower recall and vice versa. The threshold of 0.8 achieves the best trade-off mAP with $K \ge 3$. 
When $K \le 2$, a smaller threshold is better. 
This is reasonable because more boxes can be leveraged. 

Along the dimension of $K$, the precision and recall improve in general, except for $\tau=0.6$ where the precision deteriorates when $K \ge 3$. 
For $\tau = 0.8$, the accuracy improves significantly from $55.0\%$ to $59.7\%$ when the number of refinements is increased from 0 to 5. 
The gradual accuracy improvement indicates that one-step knowledge transfer is sub-optimal, and the final accuracy benefits a lot from more iterations of knowledge transfer. 

\begin{figure}[tb]
\centering
\small
\begin{tabular}{c@{}c@{}c}
\includegraphics[width=0.33\linewidth]{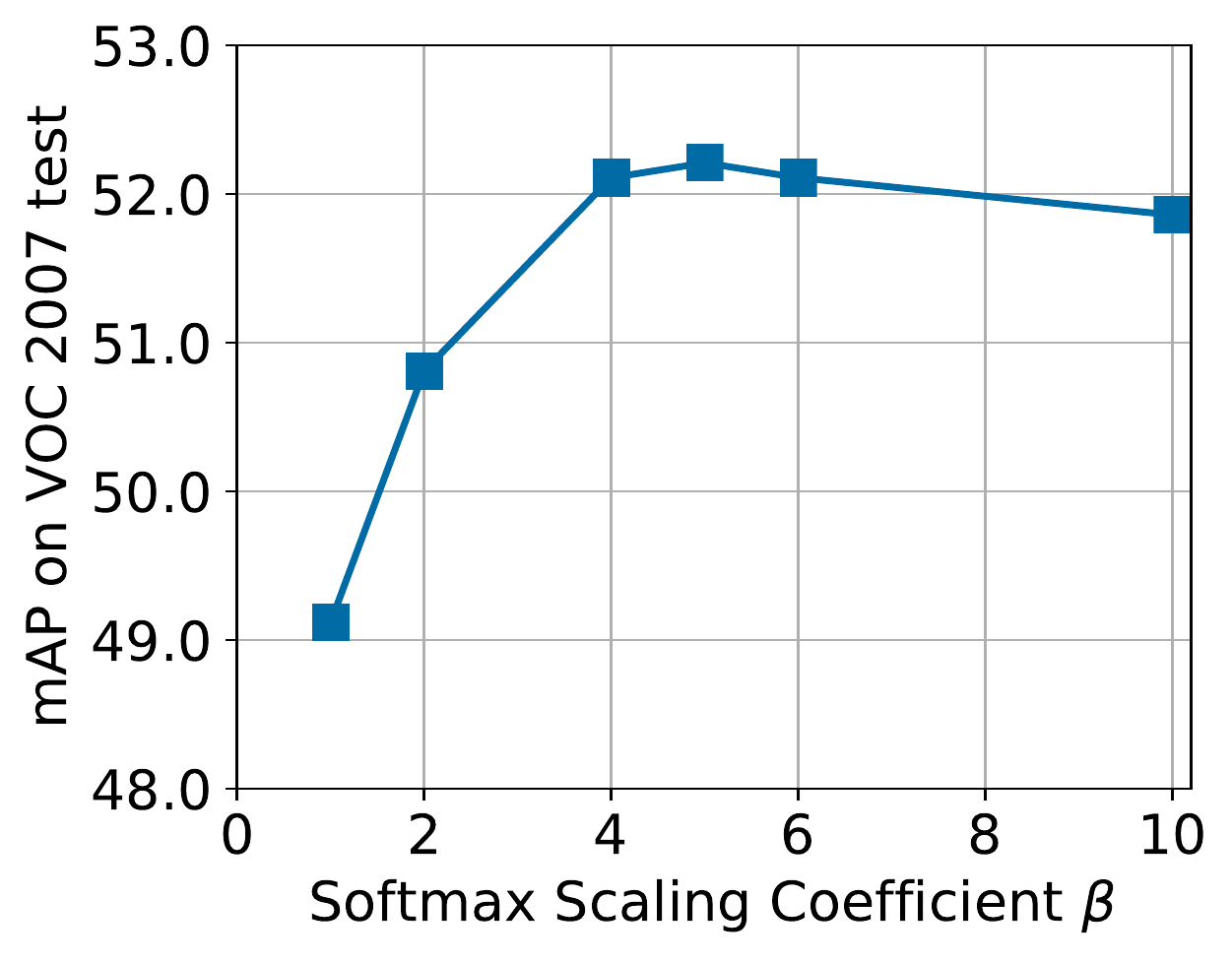}
&  
\includegraphics[width=0.33\linewidth]{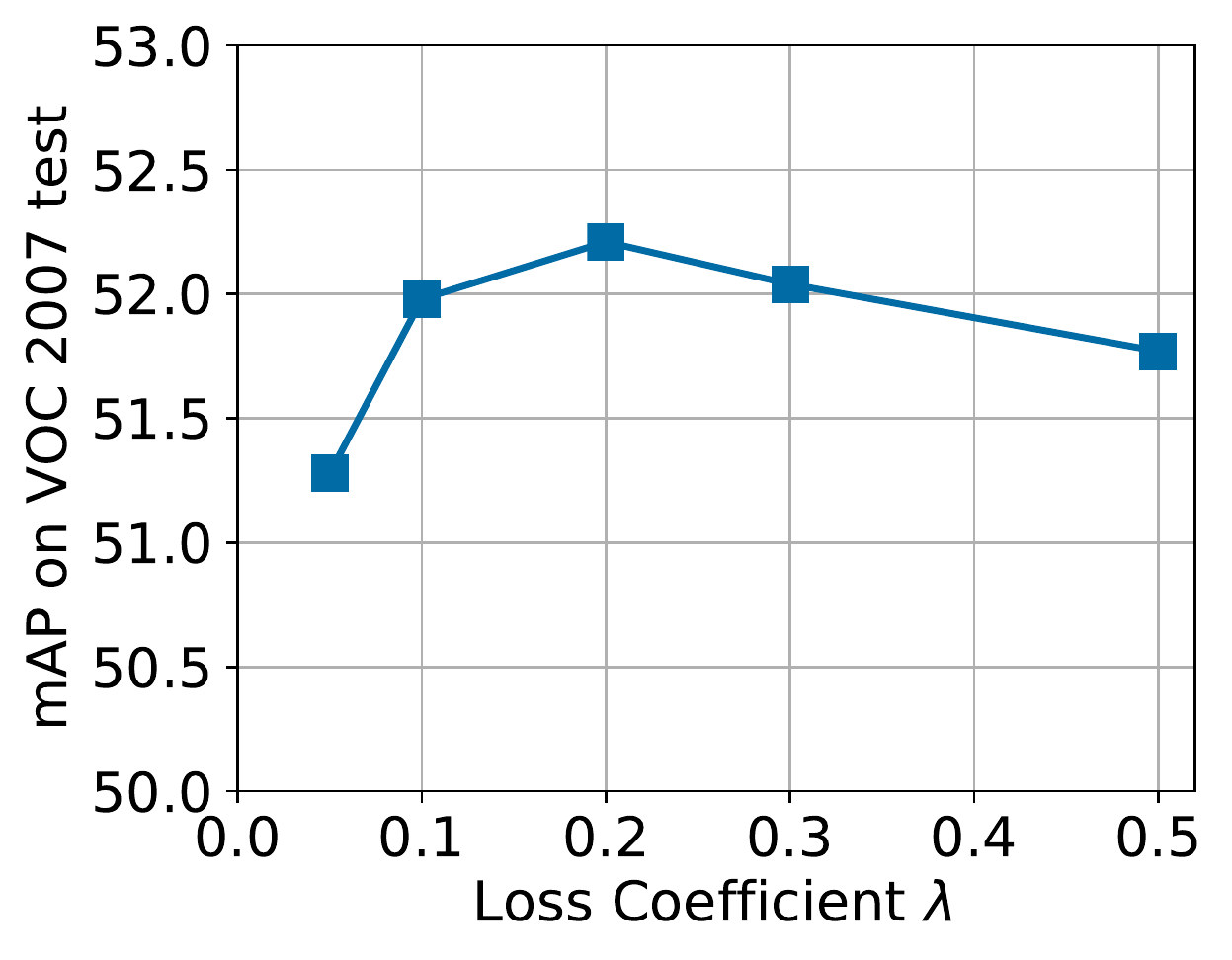} 
&
\includegraphics[width=0.33\linewidth]{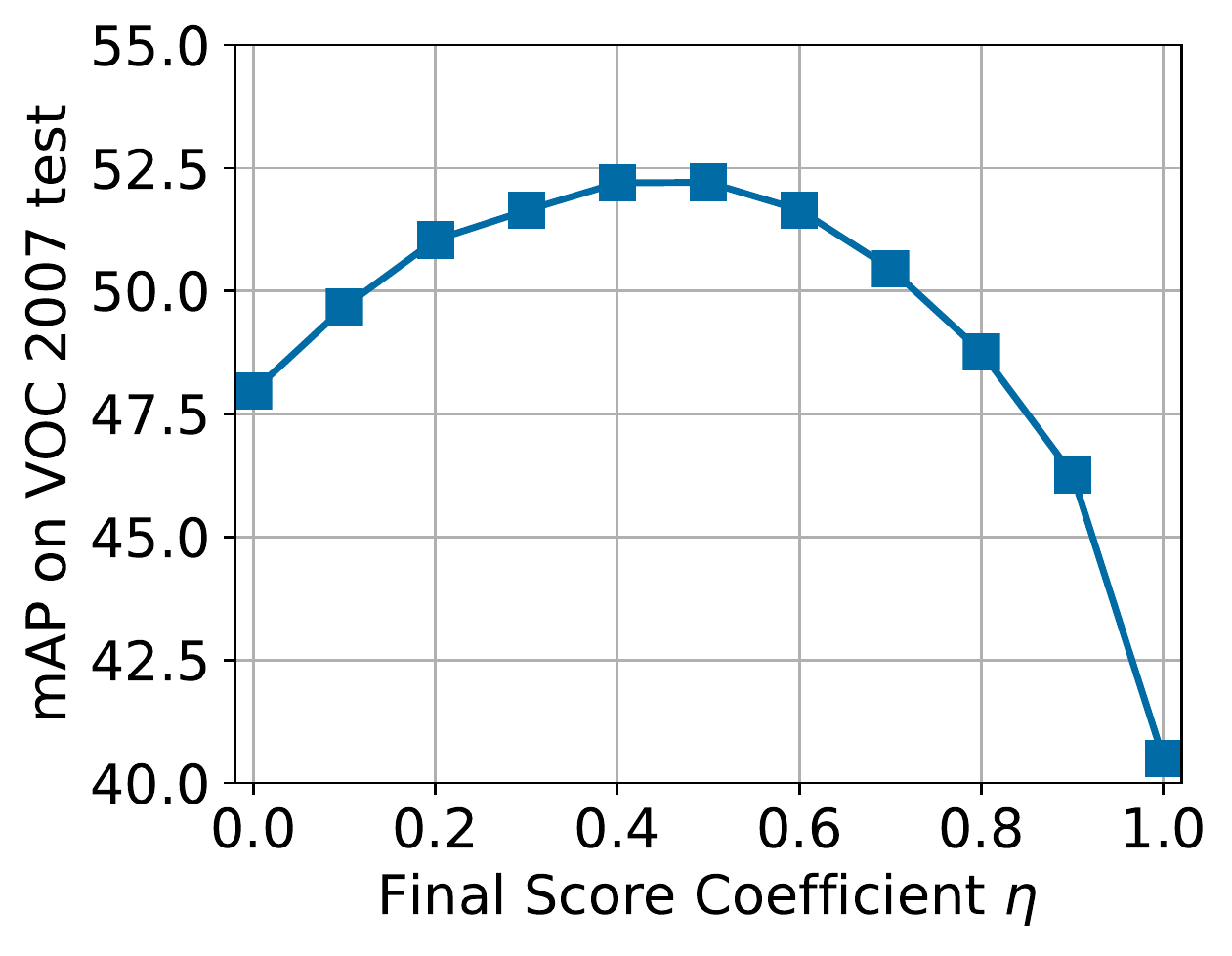} \\
(a) & (b) & (c) \\
\end{tabular}
\caption{Ablation study of the scaling factor $\beta$ in Eq.~\ref{eq:softmax_score}, $\lambda$ in Eq.~\ref{eq:loss} and $\eta$ in Eq.~\ref{eq:final_score}. The accuracy is based on the initial OCUD and MIL classifier with single-scale inference.}
\label{fig:beta}
\end{figure}

\para{$\beta$.}
The $\beta$ parameter in Eq.~\ref{fig:beta} scales the detection score $s_{ij}^d \in (0, 1)$ before softmax. When $\beta=0$, it is equivalent to remove the detection branch. When $\beta \rightarrow+\infty$, all the non-maximum values are zero after softmax, which reduces the importance of the classification branch. The best accuracy locates at $\beta = 5$ in Fig.~\ref{fig:beta}(a).

\para{$\lambda$.}
Coefficient $\lambda$ balances the image classification loss $L_{\text{wsddn}}$ and detection score regularization $L_{\text{guide}}$ in Eq.~\ref{eq:loss}. 
A larger $\lambda$ means stronger regularization. 
The result is shown in Fig.~\ref{fig:beta}(b), and $\lambda=0.2$ delivers the best performance.
A non-zero $\lambda$ performing well suggests that the OCUD can provide valuable information to guide the MIL classifier learning, which is overlooked in previous work \cite{uijlings2018revisiting}.

\para{$\eta$.}
Linear coefficient $\eta$ in Eq.~\ref{eq:final_score} balances the score from the MIL classifier and the OCUD during inference.
As illustrated in Fig.~\ref{fig:beta}(c), the accuracy is worse if we rely on either MIL classifier ($\eta = 1$) or the OCUD ($\eta = 0$) alone. The best accuracy is located at $\eta = 0.4\sim 0.5$.

\begin{figure}[t]
    \centering
    \begin{tabular}{cc}
        \includegraphics[width=0.49\linewidth]{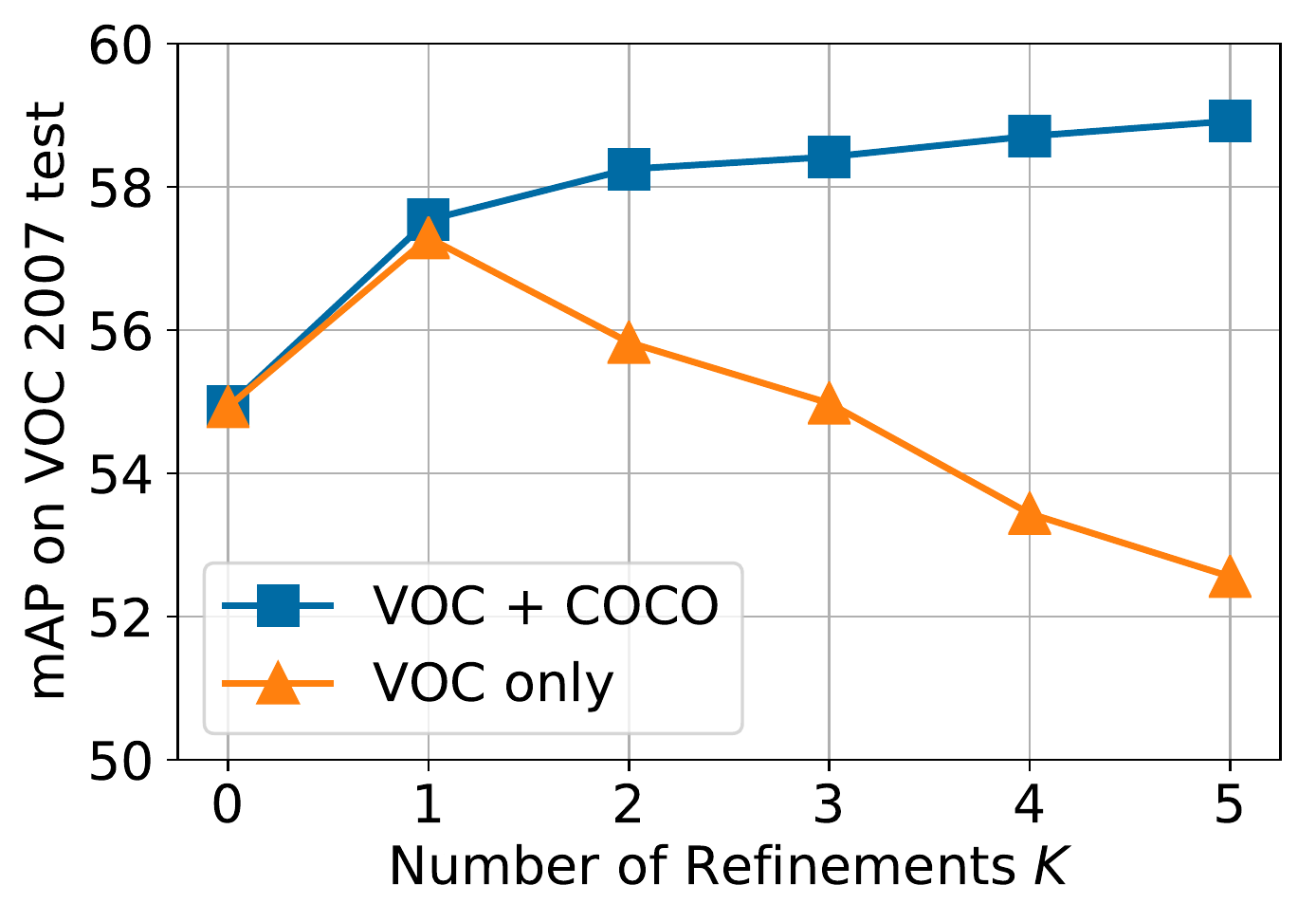} &
        \includegraphics[width=0.49\linewidth]{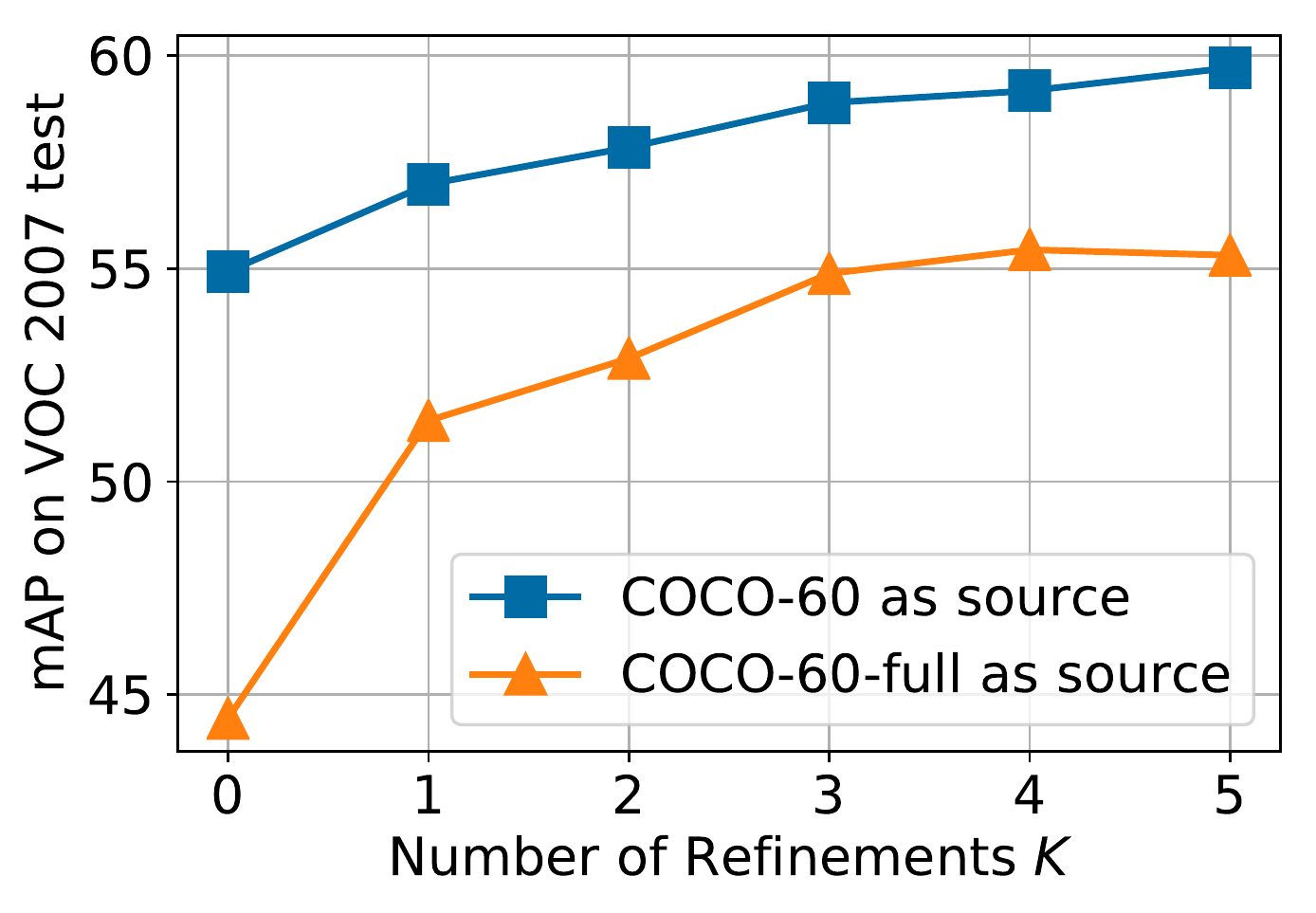}
        \\
        (a) & (b)
    \end{tabular}
    \caption{Accuracy with different configurations of the source datasets: (a) VOC+COCO vs VOC, (b) COCO-60 vs COCO-60-full. The accuracy is based on multi-scale testing.}
    \label{fig:source}
\end{figure}

\para{VOC vs VOC+COCO.}
After we have the initial OCUD, one alternative is to remove the source dataset afterwards.
Fig.~\ref{fig:source}(a) shows the experiment results with $\tau=0.7$.
As we can see, without the source dataset, the performance drops dramatically after one refinement.
The reason might be that the error of the mined box annotation can be accumulated and the OCUD becomes unstable without the guidance of the manually-labeled boxes. 
This ablation is similar to pure WSOD methods such as OICR \cite{tang2017multiple}, where the detector is refined only on the target data. The inferior result suggests that transferring knowledge from the source is indeed critical in the success of our method.

\para{COCO-60 vs COCO-60-full.}
Following \cite{lee2018universal,zhang2018mixed}, we removed all images in the source dataset (COCO) which has overlapping categories with the target VOC dataset. 
Instead of removing the images, we also conduct the experiments by keeping the images but removing the annotations of overlapping categories, and denote this source set by COCO-60-full.
Fig.~\ref{fig:source}(b) shows the experiment results.
Obviously, the accuracy with COCO-60 is higher than that with COCO-60-full.
The reason is that the regions with the annotation removed are treated as background in OCUD, which will reduce the recall rate for COCO-60-full.
Another observation is that even with this challenging source dataset, we can still boost the accuracy from less than 45\% to more than 55\%, with a gain of more than 10 points with our progressive transfer learning. 
Comparatively, the gain on COCO-60 is much less at around 5 points. 
The reason is that the propagation on the COCO-60-full can provide more positive pseudo ground truth boxes. 

\begin{figure}[tb]
\centering
\begin{tabular}{c@{~}c}
\includegraphics[width=0.49\linewidth]{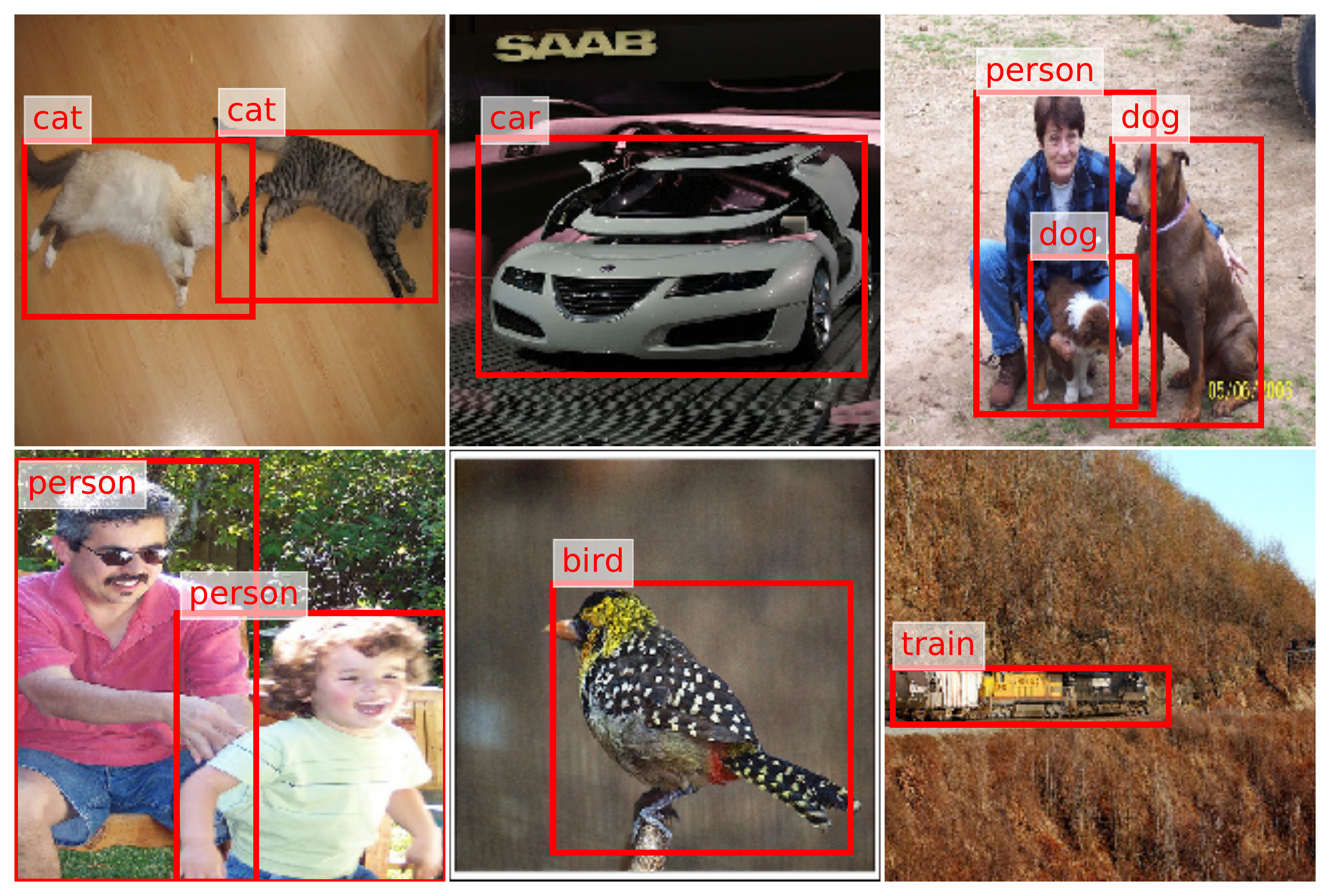}
&  
\includegraphics[width=0.49\linewidth]{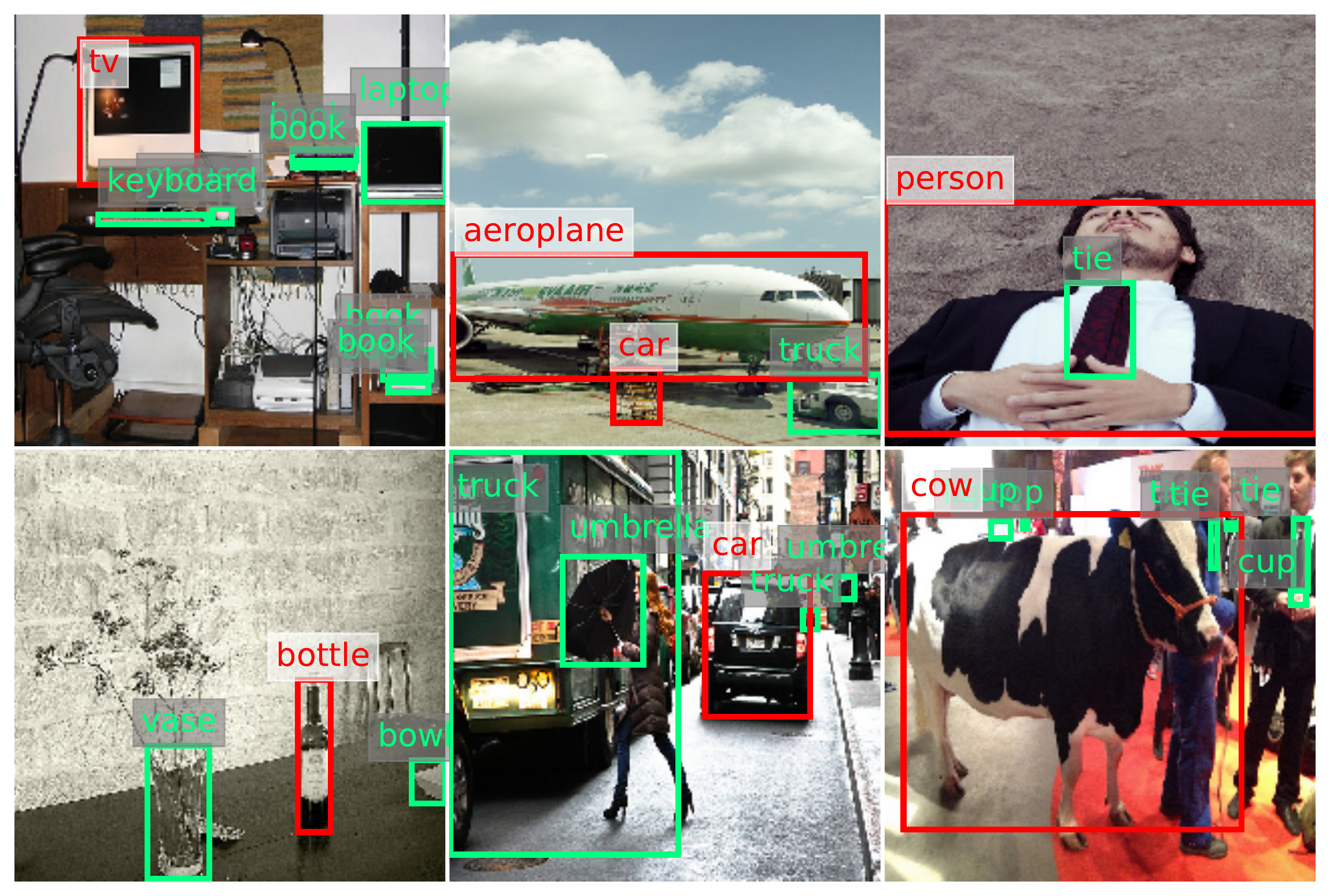}
\\
(a) & (b) \\
\end{tabular}
\caption{(a) Mined pseudo ground truth boxes (in red) in VOC trainval. (b) Original ground truth (in green) and pseudo ground truth boxes (in red) in COCO-60-full. }
\label{fig:viz}
\end{figure}

Fig.~\ref{fig:viz}(a) and \ref{fig:viz}(b) visualize the mined pseudo ground truth boxes (in red) of a few example images in the VOC trainval data and COCO-60-full after 2 refinements. From the results, we can see that some missing box-level annotations in VOC are successfully recovered, which helps the OCUD align with the target domain. 
The mined boxes in the COCO-60-full can also reduce the impact of the missing labels and improve the recall.

\begin{figure}[tb]
    \centering
    \includegraphics[width=0.98\linewidth]{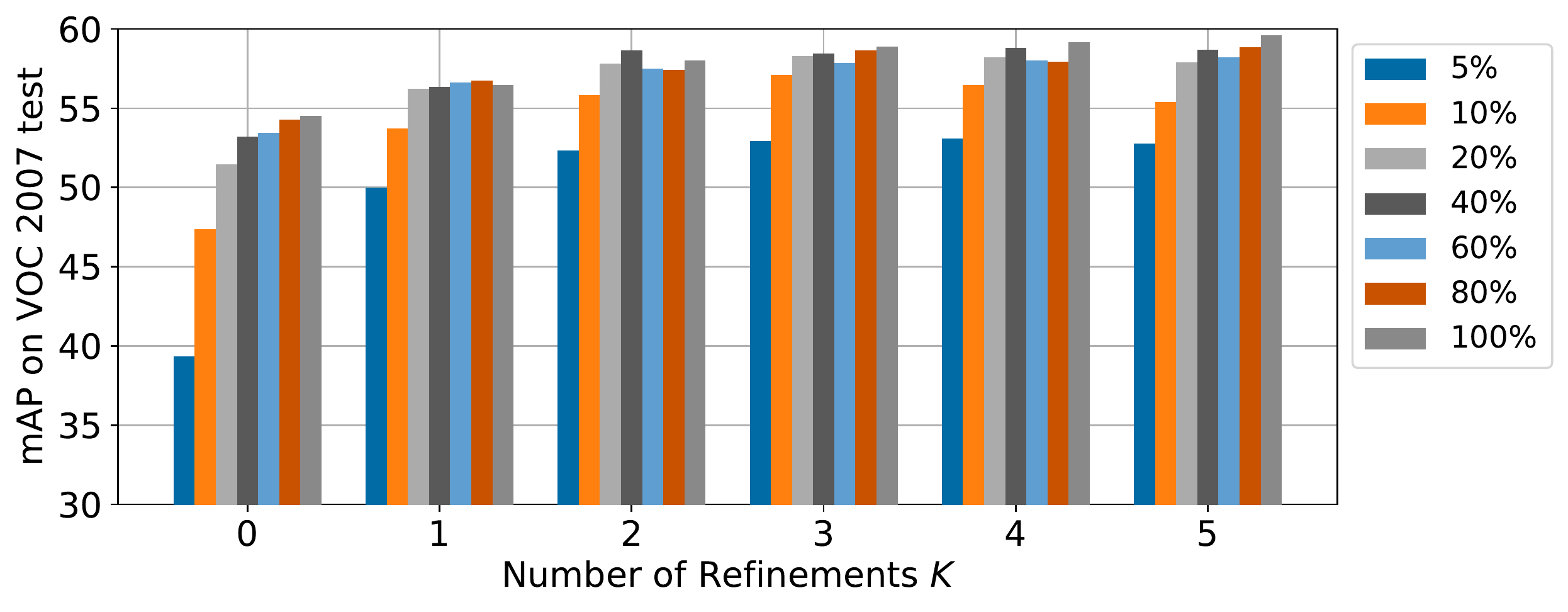}
    \caption{Ablation study on the size of the source dataset (subsets of COCO-60).}
    \label{fig:coco_percentbar}
\end{figure}

\para{Size of the Source Data.}
We study the effect of varying the source dataset's size to explore the boundary of the amount of data necessary for a successful transfer. 
Specifically, we randomly sample 5\%, 10\%, \ldots, 80\% of the COCO-60
as the source dataset. The smaller percentage subset is subsequently included in the larger percentage subset. Fig.~\ref{fig:coco_percentbar} shows the experiment results. We can observe that as few as 20\% of COCO-60 (4396 train + 194 val images) brings accuracy to more than $58\%$ mAP on VOC.

\para{COCO vs ILSVRC.}
We replace COCO-60 by ILSVRC-179 and run our algorithm for 4 refinements with the same hyper-parameters as in the COCO-60 experiment. 
The OCUD is trained with 4 times more gradient steps, because of the larger data size.
The final accuracy is 56.46\%, which is higher than COCO-60-full, but worse than COCO-60.
Compared with COCO-60-full, the superiority might come from the larger dataset. 
Compared with COCO-60, we believe the inferiority is from the data quality and consistency with the target dataset. 
Although ILSVRC-179 contains more images than COCO-60, the quality is not as good, and we observed more images with missing labels. Visual images are shown in the supplementary materials. 
This introduces more regions that are target domain objects but are taken as negative regions for OCUD.

\subsection{ILSVRC Transfer Setting}
Following the setting in \cite{hoffman2014lsda,tang2017visual,uijlings2018revisiting,zhang2018mixed}, we also conduct experiments with the 200 classes in the ILSVRC 2013 detection dataset \cite{imagenet_cvpr09}.
The setting uses the first 100 classes sorted in alphabetic order as the source classes, and the last 100 classes as the target weak classes.
We were able to achieve $37.0\%$ mAP on the weak 100 categories of val2 set with our algorithm and the ResNet50 backbone, which is comparable to the $36.9\%$ mAP reported in \cite{uijlings2018revisiting} with the stronger Inception-ResNet and is much better than any earlier results under the same setting \cite{hoffman2014lsda,tang2017visual,zhang2018mixed}.
Note that our method without any refinement is $33.5\%$, and iterative knowledge transfer boosts the performance by $3.5$ points after two refinements.
This again confirms our argument that the multi-step transfer is more effective than one-step.
The detail is provided in the supplementary material.

\section{Conclusion}

We have studied the weakly supervised object detection problem by transfer learning from a fully annotated source dataset. A simple yet effective progressive knowledge transfer algorithm is developed to learn a one-class universal detector and a MIL classifier iteratively. As such, the source dataset's knowledge can be thoroughly exploited and leveraged, leading to a new state-of-the-art on VOC 2007 with COCO-60 as the source dataset. The results suggest that knowledge transfer from an existing well-annotated dataset could be a fruitful future direction towards mitigating the annotation effort problem for novel domains.

\clearpage
\bibliographystyle{splncs04}
\bibliography{main}

\appendix
\clearpage

\section{Supplementary}

\subsection{Training and Testing Time}

One concern about our multi-step transfer framework is the elongated training time over a one-step transfer. However, we want to note that the refinements do not necessarily require a full training schedule. In the $K$-th refinement, the model is initialized from the $(K-1)$-th step, and trained with only a fractional number of steps of the initial schedule for both the one-class detector and the MIL classifier, as described in Sec.~\ref{sec:expr}.

We list the actual time in Table~\ref{tab:time}. To reach $58\%$ mAP with 1 initial training and 2 refinements, the total time is 190+23+(22+5+50+8)*2=383m, which is only 1.8 times of that of the initial model (190+23m) instead of 3 times. To reach the best $59.7\%$ mAP, 1 initial training and 5 refinements took 638m, which is 3 rather than 6 times. As progressive knowledge transfer is an offline process, we hope the higher detection mAP can pay off the longer offline cost.

\begin{table}[hb]
    \caption{Time of different training stages of the COCO-60-to-VOC experiment on a 4 Nvidia V100 GPU server.}
    \centering
    \begin{tabular}{lcc}
        \toprule
        Stage & Training steps & Wall clock time \\
        \midrule
        OCUD initial ($K=0$) & 17500 &  190 min \\
        OCUD refine ($K>0$)  & 5000  &  50 min \\
        MIL initial ($K=0$)  & 5000  &  23 min \\
        MIL refine ($K>0$)   & 2000  &  8 min \\
        Mine pseudo GT on COCO-60 & (22K images) & 22 min \\
        Mine pseudo GT on VOC trainval & (5K images) & 5 min \\
        \bottomrule
    \end{tabular}
    \label{tab:time}
\end{table}

As for the testing time, we provided optional distillation post-processing step, reported as “Ours(distill)” in the main text. The distillation retrains an ordinary Faster RCNN with the mined pseudo GT for 10000 steps in 121min, whose testing time is comparable to a usual Faster RCNN (about 2min on 4952 VOC test images). Without distillation, the combined OCUD and classifier takes roughly 5min (multi-scale).

\subsection{ILSVRC Transfer Setting}
Some previous works \cite{hoffman2014lsda,tang2017visual,uijlings2018revisiting,zhang2018mixed} have studied the transfer learning based weakly supervised object detection problem with the ILSVRC 2013 dataset \cite{imagenet_cvpr09}. We test the effectiveness of our algorithm under this setting with minimal modification from the COCO-to-VOC experiment.

\para{Source and Target Datasets.}
The ILSVRC detection dataset contains 200 object categories \cite{imagenet_cvpr09}.
We construct the source and target data following the protocol in \cite{hoffman2014lsda,tang2017visual,uijlings2018revisiting,zhang2018mixed}.
The category names are sorted in alphabetic order. The first half of 100 categories are the source domain categories, and the rest half are the target domain categories.

The val1 and val2 validation set splits as in \cite{girshick2015fast} are adopted. The val1 set originally contains 9886 images. As the source dataset, we keep 4487 images (6881 box annotations) of the first 100 categories from val1, and augment it with images sampled from the ILSVRC training set. A maximum of 1000 images per category are sampled, resulting in 78486 images (108591 box annotations).
As for the target dataset, we similarly augment the val1 images of the latter 100 categories (3609 images and 7103 annotations) with maximum 1000 ILSVRC training images (76847 images and 124991 annotations) as the target training set, while keeping only image-level labels (the box annotations are ignored). The val2 dataset with annotations of the latter 100 categories (9917 images and 14079 boxes) is used as the target test set.

\para{Evaluation Metric.}
We report the same mAP (calculated by the VOC 07 method) at IoU threshold 0.5 as in Sec.~\ref{sec:expr} of the main text on the val2 set.

\para{Network.}
ResNet-50 is used as the backbone for both the OCUD and the MIL classifier. The output channels are modified to 101 to account for the different number of classes.

\para{Training and Inference.}
Due to the increased size of the dataset, the iteration 0 OCUD was trained for 70000 steps (with a batch size of 8 images on 4 GPUs). The later refinements of OCUD took 20000 gradient steps. The MIL classifier was initially trained for 40000 steps and fine-tuned for 10000 steps during refinements. The learning rate schedule is the same as the COCO-to-VOC experiments. 
The hyper-parameters were also comparable $\beta=5, \lambda=0.2, \tau=0.8, o=0.1$, imagesize = 640, except for the larger score interpolation coefficient $\eta$: $\eta=0.85$ and the smaller prediction score threshold $0.001$.
We ran 2 refinement iterations in total.
During inference, we did single-scale testing as it delivers roughly the same performance as the multi-scale counter-part.

\begin{table}[ht]
    \caption{mAP (\%) on the ILSVRC2013 val2 set (weak 100 categories)}
    \centering
    \begin{tabular}{l c c}
        \toprule
        Method & mAP\textsubscript{.5} \\
        \midrule
        LSDA \cite{hoffman2014lsda} (AlexNet)  &  18.08 \\
        LSDA reproduced by \cite{tang2017visual} (VGG16)  &  18.86 \\
        LSDA reproduced by \cite{tang2017visual} (VGG19)  &  21.02 \\
        Tang et al. \cite{tang2017visual} (RCNN, VGG16)  &  24.91 \\
        Tang et al. \cite{tang2017visual} (RCNN, ResNet50)  &  28.30 \\
        Tang et al. \cite{tang2017visual} (Fast RCNN, VGG16)  &  26.22 \\
        Tang et al. \cite{tang2017visual} (Fast RCNN, ResNet50)  &  29.71 \\
        MSD \cite{zhang2018mixed} (AlexNet)  & 22.28 \\
        MSD \cite{zhang2018mixed} (VGG16)  & 25.26 \\
        Uijlings et al.\cite{uijlings2018revisiting} (AlexNet)  & 23.3 \\
        Uijlings et al.\cite{uijlings2018revisiting} (Inception-ResNet)  & 36.9 \\
        \midrule
        Ours(K=0, ResNet50, no augmentation)  & 32.97 \\
        Ours(K=0, ResNet50, flip)  & 33.50 \\
        Ours(K=2, ResNet50, no augmentation) &  36.90 \\
        Ours(K=2, ResNet50, flip) &
        37.00 
        \\
        \bottomrule
    \end{tabular}
    \label{tab:ilsvrc}
\end{table}

\para{Results.}
We compare results from the ILSVRC transfer experiment with prior methods in Table~\ref{tab:ilsvrc}. Our initial iteration detector without refinement (Ours(K=0)) achieved 33.50\% mAP. After 2 refinements, the mAP was boosted to 37.00\% (Ours(K=2, ResNet50, flip)).
Our result is on-par with the best previously-known result under the same protocol \cite{uijlings2018revisiting} and better than most existing works. However, there was a performance gap between our initial detector and the one in \cite{uijlings2018revisiting}, although these two should be comparable due to the method similarity. We suspect that the reason behind this is the use of a weaker backbone in our experiment (ResNet-50) compared to the Inception ResNet. We are currently investigating deeper into this.
Nevertheless, the improved performance after refinements again demonstrates the effectiveness of our progressive knowledge transfer method.

\subsection{More Visualizations}
We visualize more images to support the arguments made in the main text.

\begin{figure}[htbp]
\centering
\includegraphics[width=\linewidth]{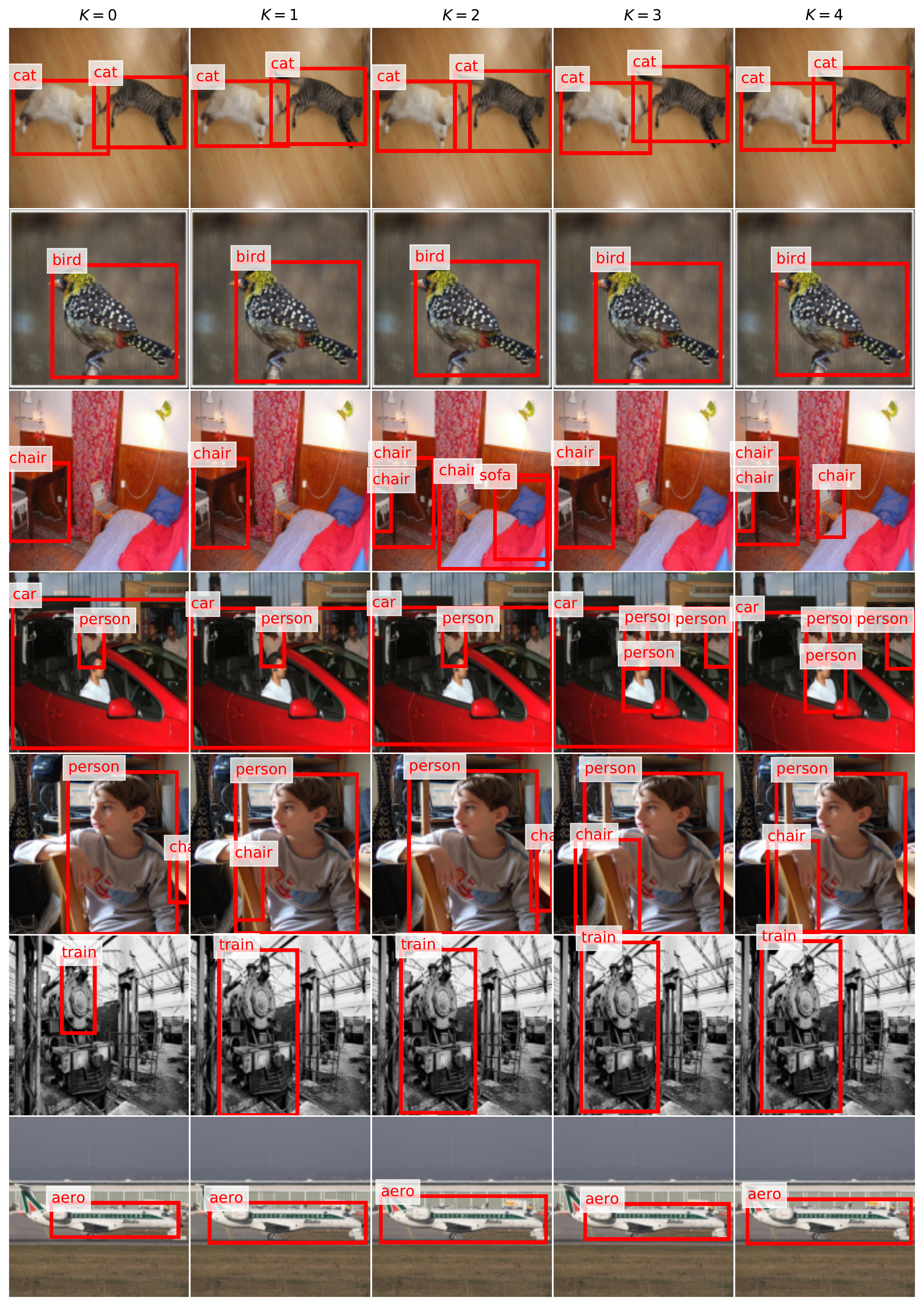}
\caption{Successfully mined pseudo ground truths (in red) in VOC trainval at number of refinements $K=0,1,2,3,4$ of the COCO-60-full to VOC experiment. }
\label{fig:more_viz_voc}
\end{figure}

\begin{figure}[htbp]
\centering
\includegraphics[width=\linewidth]{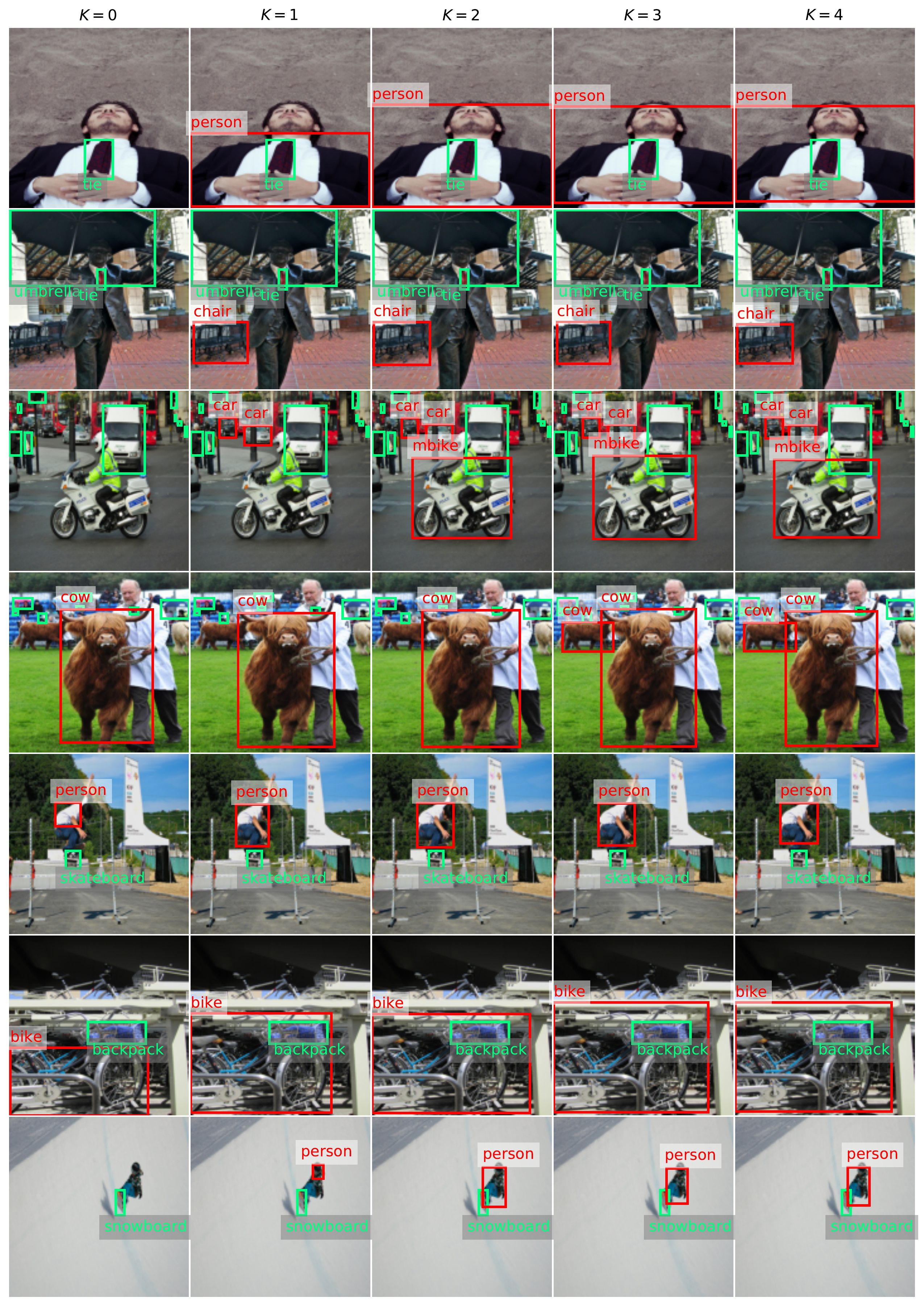}
\caption{Successfully mined pseudo ground truths (in red) in COCO-60-full at number of refinements $K=0,1,2,3,4$ of the COCO-60-full to VOC experiment. }
\label{fig:more_viz_coco}
\end{figure}

\para{COCO-60-full to VOC Transfer.}
In the main text, we visualize the mined pseudo ground truth after 2 refinements in Fig.~\ref{fig:viz}. More successful examples are shown in Fig.~\ref{fig:more_viz_voc} and \ref{fig:more_viz_coco}. We further observed that more refinement iterations could sometimes lead to higher quality pseudo ground truths by, for example, discovering objects missed by previous iterations or localizing objects more accurately.

\begin{figure}[htbp]
\centering
\includegraphics[width=\linewidth]{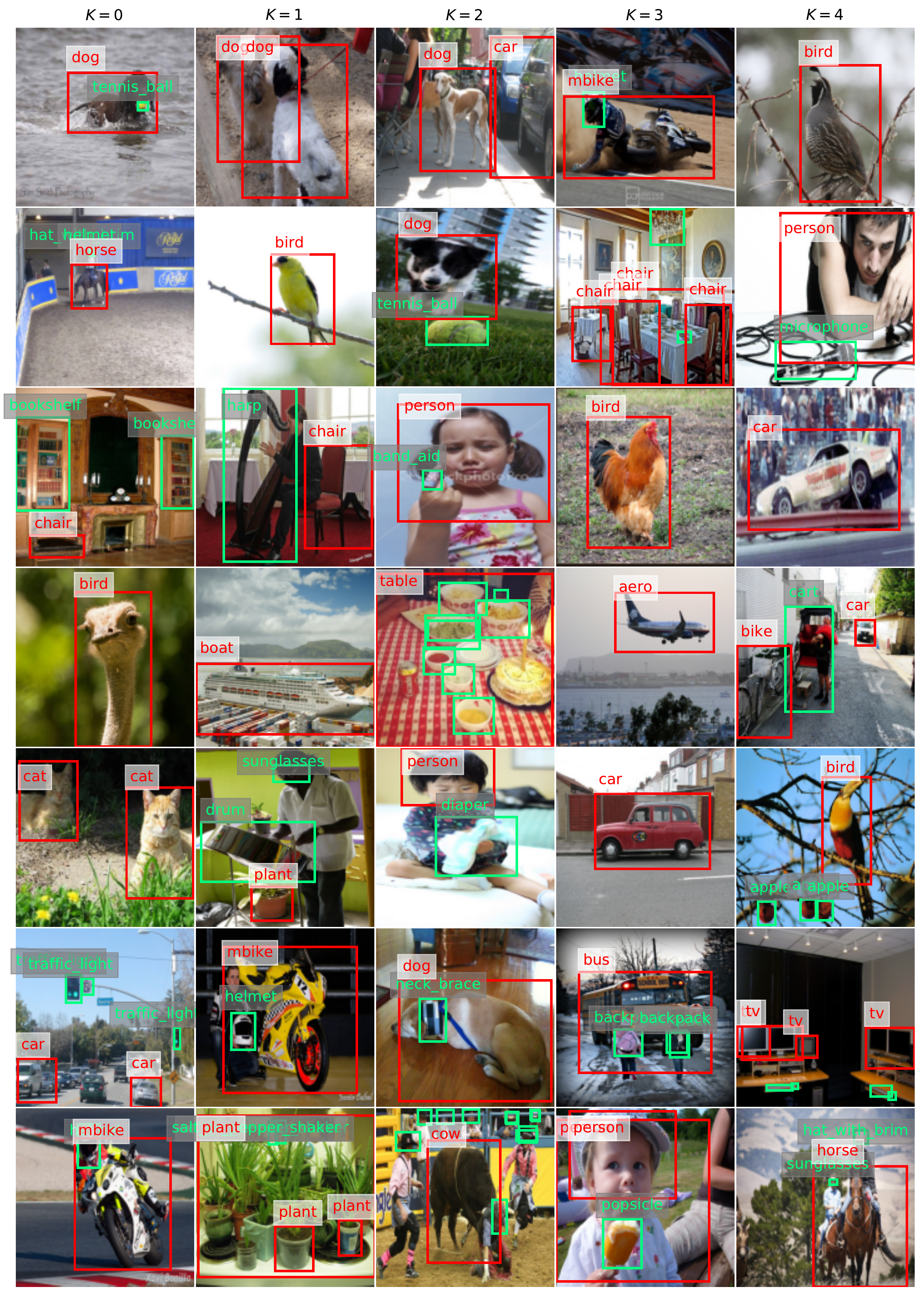}
\caption{Successfully mined pseudo ground truths (in red) in ILSVRC-179 at different iterations of the ILSVRC-179 to VOC experiment. }
\label{fig:more_viz_ilsvrc}
\end{figure}

\para{ILSVRC-179 to VOC Transfer.}
Compared with COCO-60 and COCO-60-full, the ILSVRC-179 has more missing annotations and less consistency with VOC. This can be seen from the basic statistics of the datasets and the visualizations (Fig.~\ref{fig:more_viz_ilsvrc}). The ILSVRC-179 has an average of $\frac{345854+55502}{395909+20121}\approx 0.96$ boxes per image while COCO-60 has $\frac{70549+2914}{21987+921}\approx 3.21$, and VOC test has $\frac{12032}{4952}\approx 2.43$.

\end{document}